\documentclass[10pt,journal,compsoc, letterpaper]{IEEEtran}
\usepackage{multirow}
\ifCLASSOPTIONcompsoc
  \usepackage[nocompress]{cite}
\else
  \usepackage{cite}
\fi

\ifCLASSINFOpdf
  \usepackage[pdftex]{graphicx}
\else
\fi

\usepackage[utf8]{inputenc}

\usepackage{booktabs}
\usepackage{times}
\usepackage{epsfig}
\usepackage{graphicx}
\usepackage{amsmath}
\usepackage{amssymb}

\usepackage{multirow}

\usepackage{adjustbox}

\usepackage[pagebackref=true,breaklinks=true,letterpaper=true,colorlinks,bookmarks=false]{hyperref}

\usepackage{xspace}
\makeatletter
\DeclareRobustCommand\onedot{\futurelet\@let@token\@onedot}
\def\@onedot{\ifx\@let@token.\else.\null\fi\xspace}

\def\eg{\emph{e.g}\onedot} 
\def\ie{\emph{i.e}\onedot}

\def\etal{\emph{et al}\onedot}
\makeatother

\newcommand{\argmin}{\operatornamewithlimits{argmin}}
\newcommand{\argmax}{\operatornamewithlimits{argmax}}

\newcommand{\new}[1]{\textcolor{black}{#1}}

\begin{document}


\title{LCR-Net++: Multi-person 2D and 3D Pose Detection  in Natural Images}

\author{Gr\'egory~Rogez,
		Philippe~Weinzaepfel,
        and~Cordelia~Schmid,~\IEEEmembership{Fellow,~IEEE}
\IEEEcompsocitemizethanks{
\IEEEcompsocthanksitem G. Rogez and C. Schmid are with Univ. Grenoble Alpes, Inria, CNRS, Grenoble INP, LJK, France. 
E-mail: firstname.lastname@inria.fr
\IEEEcompsocthanksitem P. Weinzaepfel is with NAVER LABS Europe, Meylan, France. \protect\\ E-mail: philippe.weinzaepfel@naverlabs.com}
}

%
%

\markboth{TO APPEAR In  IEEE TRANSACTIONS ON PATTERN ANALYSIS AND MACHINE INTELLIGENCE, 2019}{}
\IEEEtitleabstractindextext{%

\begin{abstract}
We propose an end-to-end architecture for joint 2D and 3D human pose
estimation in natural images. Key to our approach is the
generation and scoring of a number of pose proposals per image, which allows us to
predict 2D and 3D
poses of multiple people simultaneously. Hence, our
approach does not require an approximate localization of the humans
for initialization.    
Our Localization-Classification-Regression architecture, named LCR-Net, contains 3 main components: 1) the pose 
proposal generator that suggests candidate poses at different locations in the image; 
2) a classifier that scores the different
pose proposals; and 3) a regressor that refines
pose proposals both in 2D and 3D. All three stages share the 
convolutional feature layers and are trained jointly. 
The final pose estimation is obtained by integrating over neighboring
pose hypotheses, which is shown to improve over a standard non maximum suppression algorithm. 
Our method recovers full-body 2D and 3D poses, hallucinating plausible body parts  when the persons are partially occluded or truncated by the image boundary.
Our approach significantly outperforms the state of the art in 3D pose estimation on Human3.6M,  a controlled environment. 
Moreover, it shows promising results on real images for both single and multi-person subsets of the MPII 2D pose benchmark and demonstrates satisfying 3D pose results even for multi-person images.

\end{abstract}

}

\maketitle

\IEEEdisplaynontitleabstractindextext

%
\IEEEpeerreviewmaketitle


\section{Introduction}
\label{sec:intro}

\begin{figure}
  \includegraphics[width=\columnwidth]{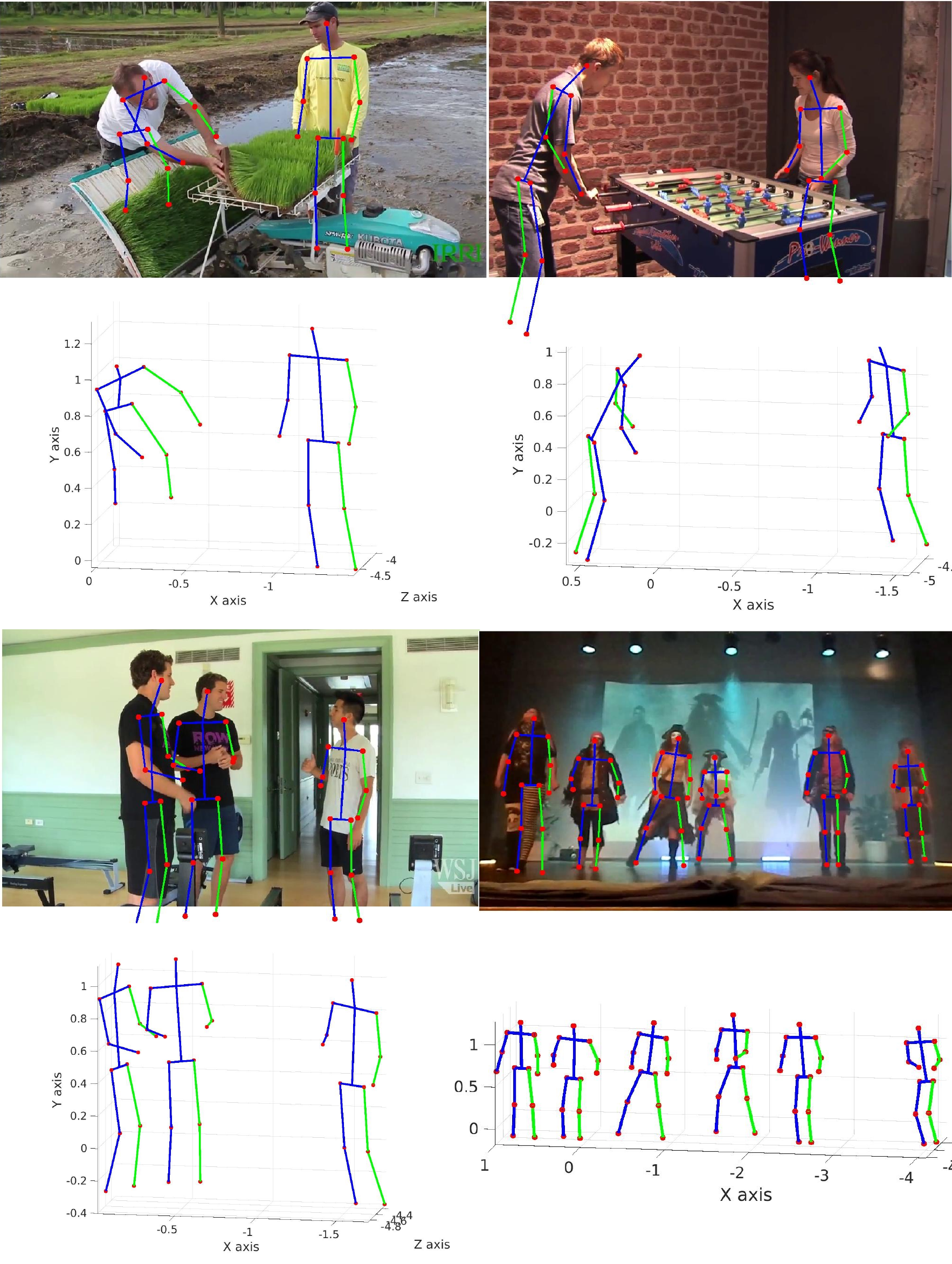}
\caption{Examples of multi-person 2D-3D pose detections in natural images. 
For each image, we show the  2D and 3D poses that are estimated jointly, even in cases of occlusions or truncations,  by reasoning in terms of full-body 2D-3D pose.}
\label{fig:2D3D}
\end{figure}

\IEEEPARstart{S}{tate}-of-the-art methods for 2D human pose estimation in real images
obtain excellent performance using Convolutional Neural Network (CNN)
architectures~\cite{CaoSWS17,BulatT16,NewellYD16}.  
However, occlusion still remains a significant challenge as analyzed in~\cite{NewellYD16}.   
One way to recover body part locations in cases of occlusions is to reason about the full-body 3D pose. 
Methods for 3D human pose understanding require training data that is only available through Motion Capture (MoCap) systems~\cite{SigalBB10,IonescuPOS14,CMUposedataset}. 
Even if they show accurate pose estimation results (including occluded joints) in controlled environments, 
these approaches do not generalize well to real images, with the
exception of recent work based on data synthesis that shows promising
results in the wild~\cite{ChenWLSW16,RogezS16}. 
In this paper, we propose a method that results in multiple full-body 2D and 3D pose hypotheses in different regions of the image. 
These {\it pose proposals} are efficiently sampled, scored and refined using an end-to-end CNN architecture inspired by the latest work on object detection~\cite{FasterRCNN}.
Finally, the pose proposals  are combined to estimate both the location and the 2D-3D pose of the individuals present in the observed scene. 
Our method recovers full-body poses, even when the persons are partially occluded or truncated by the image boundary as illustrated in the examples presented in Figure~\ref{fig:2D3D}.
\begin{figure*}
 \includegraphics[width=\linewidth]{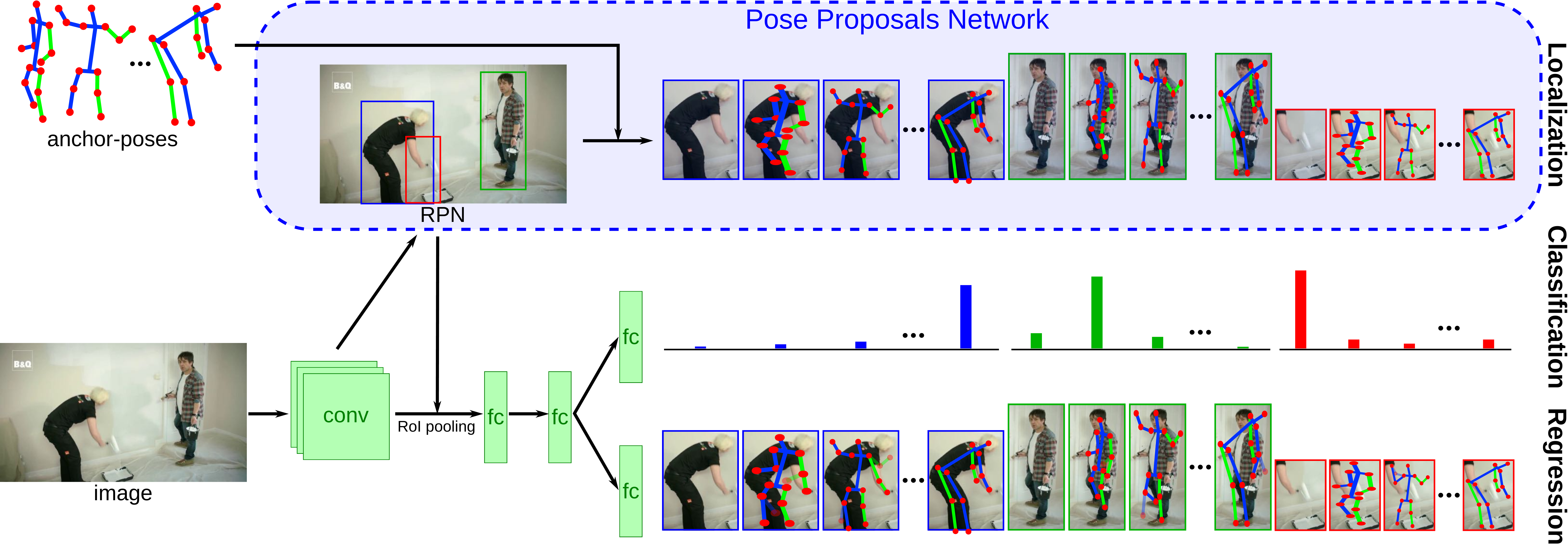}
 \caption{Overview of our LCR-Net architecture (poses only shown in 2D for better readability). We first extract candidate regions using a Region Proposal Network (RPN) and obtain pose proposals by placing a fixed set of anchor-poses into these boxes (top). These pose proposals are then scored by a classification branch
 and refined using class-specific regressors, learned independently for each anchor-pose.}
 \label{fig:archi}
\end{figure*}

CNNs have been used for full-body pose estimation both in regression~\cite{ParkHK16,TekinRLF16,ChenWLSW16,LiC14,ToshevS14_DeepPose} and classification~\cite{RogezS16} approaches. 
Regression networks are trained to directly estimate the 2D or 3D
location of the body joints, whereas a classification approach defines pose classes and returns the average pose of the top scoring class. 
Increasing the number of clusters improves precision of the estimation in classification approaches but makes discrimination harder. 
Regression methods can only predict one pose for a given image and fail to model multi-modal outputs, e.g., for ambiguous cases. 
In this paper, we argue that for full-body human pose estimation, the discriminative power of classification networks can be combined with the smoothness of regression methods 
by a simple yet elegant modification within the learning procedure.
The architecture is similar in spirit to Faster R-CNN~\cite{FasterRCNN}
which jointly localizes and classifies objects while regressing a refined bounding box.
This is achieved using a Region Proposal Network (RPN) that generates high-quality region proposals where object bounds and objectness scores are predicted. Instead of classifying objects, we propose to classify human poses.
The key idea of our approach is to quantify  the space of valid full-body poses and jointly train a K-way classifier on this partitioned space as well as local pose regression models, e.g. one per pose cluster.
To this end, we formulate a joint classification-regression loss function that combines coarse pose classification and class-specific pose regression. 
Given a set of K hypothetical pose classes, we output for each proposed image region a list of K refined 2D-3D poses and the associated classification scores.

In summary, we propose an end-to-end  Localization-Classification-Regression architecture, named LCR-Net, that detects 2D and 3D poses in natural images, see Figure~\ref{fig:archi}.
The network proceeds by extracting candidate regions for the person localization.
We obtain \emph{pose proposals} by locating the set of K hypothetical pose classes, denoted as anchor-poses, in these candidate boxes.
Each pose proposal is then scored using a classification branch and regressed independently for each anchor-pose.
The localization, \ie, extraction of the pose proposals, classification and per anchor-pose regression, share layers and can be trained end-to-end.
Our final output consists in a number of
2D-3D poses per images that are obtained by aggregating similar pose proposals, in
terms of 2D location and 3D pose. 
To the best of our knowledge, our work is the first to tackle multi-person 3D pose estimation from a single image.
The work presented in this paper is an extension of~\cite{RogezWS17}. We analyze four ways to improve the 2D-3D pose estimation performance of our LCR-Net architecture: (1) the use of additional synthetic data to augment the size of the training data sets, (2) a variant of the architecture with an iterative process as in~\cite{NewellYD16,CaoSWS17} that further refines regression and classification results, (3) an improved alignment of the candidate regions of interest as in~\cite{he2017mask} that better conserves spatial details by avoiding rounding operations \new{and 4) a ResNet~\cite{HeZRS16} backbone to increase learning capacity}. Altogether, this version referred to as LCR-Net++, significantly improves over the initial version, with a boost in performance of more than 20mm in 3D and \new{10\% in 2D pose accuracy (PCKh@0.5)}. Our approach outperforms the state of the art for 3D pose estimation in a controlled environment, even when compared to methods that leverage temporal smoothing or rely on initial localization of the human. It shows promising results for real images, estimating poses in  2D and 3D.

After reviewing the related work in Section~\ref{sec:related}, Section~\ref{sec:method} introduces LCR-Net and its variants.
Extensive experimental results, both in 2D and 3D, are presented in Section~\ref{sec:xp}.

\section{Related work}
\label{sec:related}

In this section, we review related work for 2D (Section~\ref{sub:related2D}) and 3D (Section~\ref{sub:related3D}) human pose estimation from single images.

\subsection{Human localization and 2D pose estimation} 
\label{sub:related2D}

Most state-of-the-art approaches for 2D human pose estimation employ CNN architectures~\cite{CaoSWS17,BulatT16,ChenY14,FanZLW15,NewellYD16,OuyangCW14,TompsonJLB14,ToshevS14_DeepPose}. 
They can be divided into two groups: (a) methods which first search the 
image for local body parts and model their dependencies using graphical models~\cite{ChenY14,TompsonJLB14,CaoSWS17}
and (b) holistic approaches that directly estimate the full body~\cite{ToshevS14_DeepPose,FanZLW15}.

Methods based on local body parts require a tight bounding box around each human to estimate his pose~\cite{WeiRKS16,NewellYD16,PapandreouZKTTB17},  others can detect multiple people in natural images at once~\cite{CaoSWS17}.
Most methods extract joint heatmaps, \ie, probabilistic maps that estimate the probability of each pixel to contain a particular joint.
An iterative procedure is often used~\cite{WeiRKS16,NewellYD16,CaoSWS17}:
a refined estimate of the heatmaps is obtained from the previous estimate and the convolutional features.
Joint positions can be estimated by taking the local maxima of the heatmaps.
In Convolutional Pose Machines, Wei \etal~\cite{WeiRKS16} refine the predictions over successive stages with intermediate supervision at each stage. 
In the Stacked Hourglass network~\cite{NewellYD16}, repeated bottom-up, top-down processing used in conjunction with intermediate supervision improve the performance of the network.

Papandreou \etal~\cite{PapandreouZKTTB17} also compute a per-joint regressor at each pixel to refine the position of the joints, that may lack precision due to the stride of CNNs.
Given the joint positions extracted from the heatmaps, additional post-processing is required to build human poses, such as graph partitioning~\cite{PishchulinITAAG15}.
Cao \etal~\cite{CaoSWS17} proposed an alternative approach by also regressing affinities between joints, \ie, the direction of the bones, together with the heatmaps.
In contrast to these methods that build human poses from local body parts, our method extract full-body 2D and 3D poses, even in case of occlusions.

Holistic approaches often assume that the individuals have been localized,
and that a bounding box around each person is available.
Toshev and Szegedy~\cite{ToshevS14_DeepPose} directly regress the positions for each joint using an iterative procedure.
Fan \etal~\cite{FanZLW15} combines the local appearance with an holistic view of the body to estimate the position of the joints.
Instead of relying on a multi-stage approach, our network is trained in an end-to-end fashion and outputs both 2D and 3D poses jointly.

\subsection{3D human pose from a single image}
\label{sub:related3D}

Methods for 3D human pose estimation from a single image can be decomposed into two groups: (a) the ones that first compute 2D poses and use them to estimate 3D poses and (b) approaches that directly learn mappings from image features to 3D poses.

Motivated by the recent advances in 2D pose detection, a large body of work tackles 3D pose estimation from 2D poses assuming that the 2D joints are available~\cite{AkhterB15,FanZZW14}, provided by an off-the-shelf 2D pose detector~\cite{BogoKLGRB2016,Simo-SerraRATM12,WangWLYG14,RamakrishnaECCV2012,ChenR17,Moreno17,NieWZ17,MartinezHRL17} or obtained through a 2D pose estimation module within the proposed architecture~\cite{IqbalGG16,LinLLWC17}. 
Most of these methods reason about geometry. 
Chen and Ramanan~\cite{ChenR17} estimate 3D pose from 2D through a simple nearest neighbor search on a given 3D pose library with a large number of 2D projections. Moreno-Noguer~\cite{Moreno17} formulates the problem as a 2D-to-3D distance matrix regression.
Nie \etal~\cite{NieWZ17} predict the depth of human joints based on their 2D locations using LSTM, whereas Martinez \etal~\cite{MartinezHRL17} lift 2D joints to 3D space using a simple, fast and lightweight deep neural network.
These methods remain limited by the performance of the 2D pose estimator.

Some other approaches directly estimate the 3D pose from image features~\cite{AgarwalCVPR2004,RogezRROT08,SminchisescuCVPR2005,BoCVPR2008,ShakhnarovichCVPR2003}. 
Recently, this has been naturally extended to end-to-end mappings using CNN architectures, either in monocular
images~\cite{LiC14,LiZC15,TekinBMVC2016,RogezS16,ChenWLSW16,PavlakosZDD17} or in videos~\cite{TekinRLF16,ZhouZLDD16}.   
Pavlakos \etal~\cite{PavlakosZDD17} propose a volumetric representation for 3D human pose and employ a ConvNet to predict per-voxel likelihoods for each joint.
In~\cite{SunSLW17}, a structure-aware regression approach is followed with a reparameterized pose representation using bones instead of joints.

Finally, some recent approaches treat  2D and 3D pose estimation jointly or iteratively~\cite{SunSLW17,Simo-SerraQTM13,ZhouT14,TekingMSF17,TomeRA17}. 
In \cite{TekingMSF17}, the authors learn how to fuse 2D and 3D image cues while in~\cite{TomeRA17} a multi-stage CNN architecture leverages the knowledge of plausible 3D landmark locations to refine the search for better 2D locations. 
Most similar to our approach is the classifier of~\cite{RogezS16} that outputs a distribution of scores over a quantized set of 2D-3D poses. We also use a classifier where each class corresponds to a particular 2D-3D orientated pose but we combine classification and regression in an effective architecture that refines the pose using a class-specific regression stage. Importantly, the method of~\cite{RogezS16} requires a well-aligned bounding box around the subject while we jointly localize and estimate 2D and 3D pose of multiple people in real-world images.

Large-scale training data is necessary to train accurate state-of-the-art CNN architectures for pose estimation.
While 2D pose data are obtained by manually annotating images captured in-the-wild, reliable 3D poses are acquired using motion capture (MoCap) systems in constrained environment.
As a consequence, many methods for 3D pose estimation are trained and evaluated in these controlled and unrealistic scenarios~\cite{SigalBB10,IonescuPOS14} and do not generalize well to real-world images. Some architectures have been proposed to take advantage of the different sources of training data, \ie, indoor images with MoCap 3D poses and real-world images with 2D annotations~\cite{MehtaRCFSXT17,ZhouHSXW17}.
To  generalize to in-the-wild images, Mehta \etal~\cite{MehtaRCFSXT17} proposed a 2D-to-3D knowledge transfer, \ie, using pre-trained 2D pose networks to initialize the 3D pose regression networks while in~\cite{ZhouHSXW17} the common representations between the 2D and the 3D tasks are shared. 
 To compensate for the lack of large scale in-the-wild datasets, recent work has also proposed to generate training images for particular 3D pose datasets such as the CMU MoCap dataset~\cite{CMUposedataset} by stitching image regions~\cite{RogezS16}, animating human 3D models~\cite{ChenWLSW16,DesouzaGCL17}, using a game engine~\cite{HuangR17} or by rendering textured 3D body scans~\cite{VarolRMMBLS17,RogezS18}. These synthetic datasets have proved to be useful for training CNN architectures, yet often requiring a domain adaptation stage. However, none is realistic enough in terms of clothing, hair or interactions with objects to be considered as a fully-convincing alternative to real images.  Recently, Lassner \etal~\cite{Lassner0KBBG17} proposed a self-improving, scalable method that obtains high-quality 3D body model fits for 2D images. We also generate ``pseudo'' ground-truth 3D pose annotations for real-world images following a simple yet effective method that leverages 2D pose annotations to 3D using large-scale motion capture data.

\section{LCR-Net}
\label{sec:method}
 
We propose to detect human poses using a Localization-Classification-Regression Network (LCR-Net).
In this paper, a human pose $(p,P)$ is defined as the 2D pose $p$, 
\ie, the pixel coordinates of each joint in the image, 
and the 3D pose $P$, \ie, 3D location of each joint relative to the body center (in meters).
We consider poses with \new{$J =$} 13 joints.
We assume that a fixed set of $K$ 2D-3D anchor-poses is given, denoted by $\{ ( {a}_k , A_k) \}_{k=1..K}$. 
In this paper, they are obtained by clustering a large set of poses and using the center of each cluster as anchor pose, see Section~\ref{sec:xp} for details.

Figure~\ref{fig:archi} shows an overview of our LCR-Net architecture.
Given an image, we first compute convolutional features.
The \emph{Localization} component, also called Pose Proposals Network in the context of pose detection,  outputs a list of pose proposals. 
Pose proposals consist of a set of candidate locations where the anchor-poses are hypothesized.
Next, a Region-of-Interest (RoI) pooling layer aggregates the features inside each candidate region.
After two fully-connected layers, the network is split into two components.
The \emph{Classification} branch estimates the probability of anchor-poses to be correct at each location.
It thus jointly learns to localize humans, as well as to estimate
which anchor-pose is more probable.
The \emph{Regression} branch computes an anchor-pose-specific
regression that estimates the difference between the true human pose and the pose proposal (Figure~\ref{fig:regression}).
Our loss is the sum of three losses that we describe in more detail in the following:
\begin{equation}
\mathcal{L} = \mathcal{L}_{Loc} + \mathcal{L}_{Classif} + \mathcal{L}_{Reg} ~~.
\label{eqn:loss}
\end{equation}

Note that the convolutional features are shared between the three components 
and that the classification and regression branches also share features from two fully-connected layers. 
The architecture allows end-to-end training for localizing humans and
estimating their 2D-3D poses, in contrast to most previous works which run a human detector before estimating the pose.

\subsection{Localization: pose proposals network}

The Pose Proposal Network outputs a set of N$\times$K pose proposals, \ie, 2D-3D pose hypotheses obtained by placing the K anchor-poses in the N bounding boxes generated by the RPN~\cite{FasterRCNN}.  
These pose proposals will be scored and refined by the classification and regression branches respectively, see Figure~\ref{fig:archi}.
The loss of the localization component is the loss of the RPN network:
\begin{equation}
\mathcal{L}_{Loc} = \mathcal{L}_{RPN} ~~.
\label{eqn:locloss}
\end{equation}

During training, each bounding box $B$ is labeled with a ground-truth class $c_B \in \{ 0 \ldots K \} $ and a pose regression target $t_{c_B}$.
The ground-truth class $c_B$ is set to $0$ (corresponding to background) if the bounding box has an Intersection over Union (IoU)
below 0.5 with all ground-truth poses.
The IoU between a box and a pose is computed using the bounding box around all joints of the pose, with a fixed additional margin of 10\%.
If $B$ has a high overlap with several poses, let $(p,P)$ be the
ground-truth pose with the highest IoU with the box. 
\new{The class label $c_B$ is set by finding the closest 3D anchor-pose $A_{c_B}$ according to the distance $D_{3D}(.,.)$ between oriented 3D poses centered at the torso: $c_B = \argmin_{k} D_{3D}(A_k, P)$.
This label will be used by the classification branch (Section~\ref{sec:classif}).
If the label $c_B$ is non-zero for a box, we also define a pose regression target $t_{c_B}$, see arrows on Figure~\ref{fig:regression}, used in the regression branch (Section~\ref{sec:regression}): $t_{c_B} = (\tilde{p}-\tilde{a}_{c_B},P-A_{c_B})$, where $\tilde{p}$ and $\tilde{a}_{c_B}$ denote the 2D pose and 2D anchor-pose for class $c_B$ normalized in the range $[0..1]$ according to the box coordinates. This normalization makes the regression independent of scale and position of the person in the image. The poses $P$ and $A_{c_B}$ being both centered at the torso and expressed in meters, 2D and 3D quantities are all expressed in an approximate [-1,1] range, allowing a simultaneous regression in 2D and 3D.}

\subsection{Classification}
\label{sec:classif}

The classification component aims at predicting the closest anchor-pose, \ie, the correct label, for each bounding box $B$.
In other words, each bounding box is assigned a probability for each
anchor-pose (and the background class). 
Let $u$ be the probability distribution estimated by the network, obtained by three fully-connected layers after RoI pooling, see Figure~\ref{fig:archi}, followed by a softmax. 
The classification loss is defined using the standard log loss of the true class:
\begin{equation}
\mathcal{L}_{Classif}(u,c_B) = - log~u ({c_B}) ~~.
\label{eqn:clsloss}
\end{equation}

\subsection{Regression}
\label{sec:regression}
\begin{figure}
 \centering
 \includegraphics[width=0.8\linewidth]{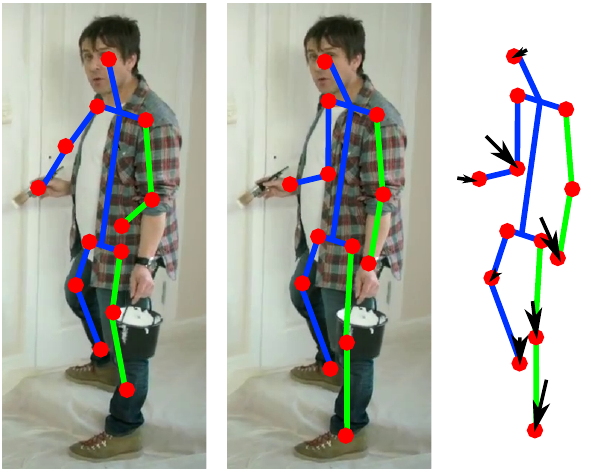}
 \caption{The regression aims at refining the anchor-pose to match the ground-truth 2D-3D pose (only shown in 2D for better readability).}
 \label{fig:regression}
\end{figure}

The regression component aims at refining the coarse anchor-poses located in the region proposals as depicted in Figure~\ref{fig:regression}.
The specificity of our approach is that the regression is anchor-pose-specific and a regressor is learned independently for each anchor-pose.
The regression outputs $v$ are obtained by using a fully-connected
layer after the two fully-connected layers shared with the classification branch (see Figure~\ref{fig:archi}).
The dimension of $v$ is equal to \new{$5 \times J \times (K+1)$}, 
where \new{$J$ is the number of joints and the factor of 5 is the coordinates (2D +3D)}. We denote by $v_{c_B}$ the subvector of $v$ corresponding to the regression for anchor-pose $c_B$.
The regression loss is defined as: 
\begin{equation}
\mathcal{L}_{Reg}(v,t_{c_B}) = [c_B \geqslant 1 ] ~~ \Vert t_{c_B} - v_{c_B} \Vert_S ~~,
\label{eqn:regloss}
\end{equation}
with $\Vert . \Vert_S$ the smooth-L1 loss, a robust version of the L2 loss which is less sensitive to outliers:
\begin{equation}
\Vert x \Vert_S = \begin{cases} 0.5 x^2 & \text{ if }\vert x \vert < 1, \\ \vert x \vert - 0.5 & \text{ otherwise. } \end{cases}
\end{equation}

\subsection{Iterative Estimation}
\label{sub:iterative}

\begin{figure}
 \centering
 \includegraphics[width=\linewidth]{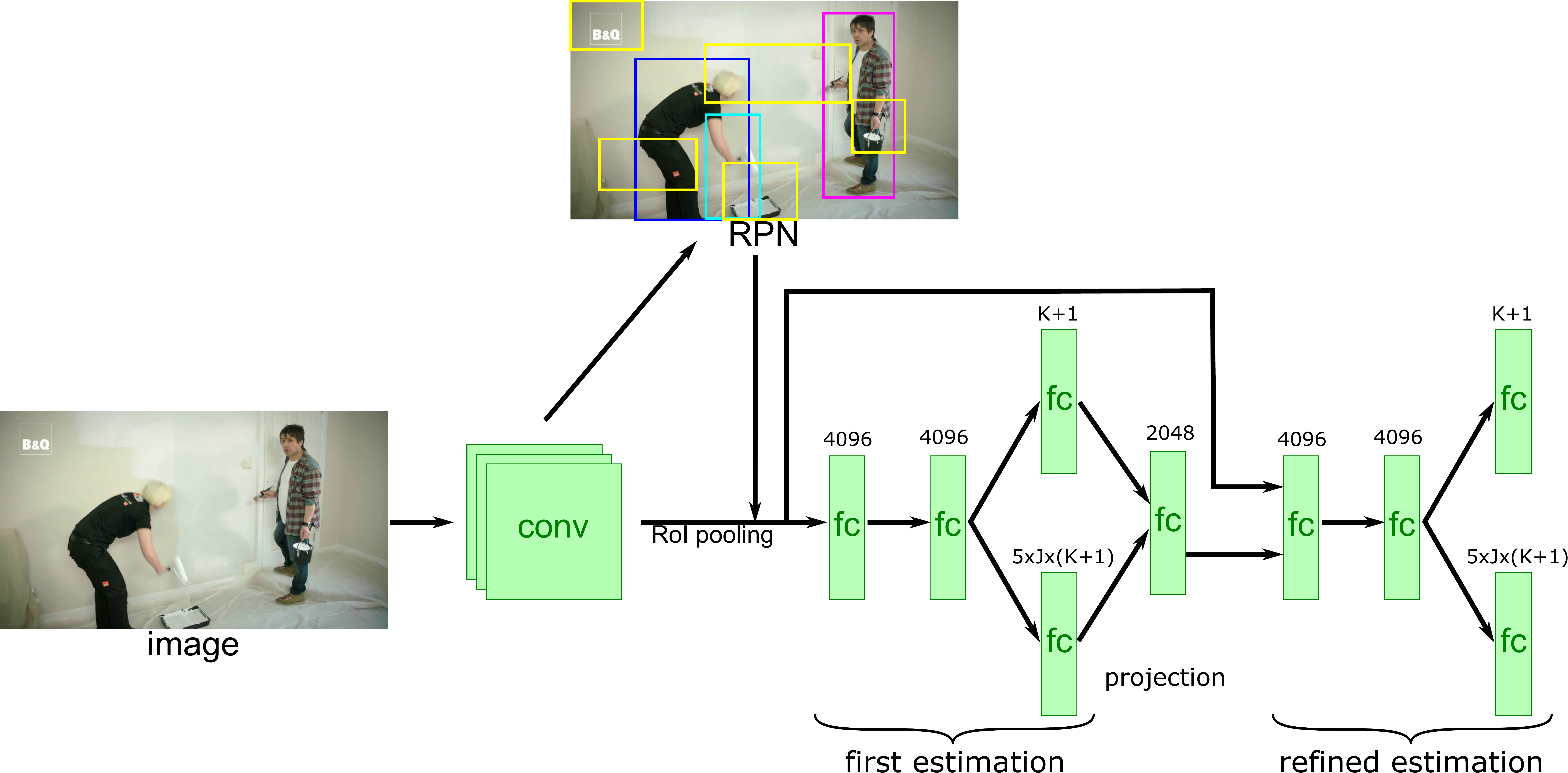}
 \caption{Illustration of the iterative estimation procedure. \new{The classification branch outputs K+1 scores, one per class plus background. The regression branch outputs $5 \times J \times (K+1)$ values, the regression being class-specific and outputting 2D and 3D values for each of the J joints.}}
 \label{fig:iterative}
\end{figure}

We propose a variant of the architecture in which the regression and classification are iteratively estimated and refined.  Such an iterative estimation is common in pose estimation~\cite{NewellYD16,CaoSWS17}. 
More precisely, we add several layers at the end of the LCR-Net networks, see Figure~\ref{fig:iterative}. A first estimate of the classification and regression is obtained using two fully-connected layers, that are shared between the two tasks, followed by a fully-connected layer for each task. The result of this first estimate is combined with the features pooled over the RoI to refine the estimation. In more details, the output of these first classification and regression are concatenated and fed to a fully-connected layer to obtain a fixed representation of 2048 dimensions, independently of K. We then concatenate this feature vector with the convolutional features pooled over the RoI and feed it to a similar network architecture as done for the initial estimate: two fully-connected layers followed by an additional layer for classification and another one for regression. Losses are applied after the first estimate and its refinements during training, while only the last estimation is returned at test time.

\subsection{Implementation details}

Similar to Faster R-CNN, we use an approximate joint training version, 
in which boxes are considered as fixed by the RoI pooling layer.
We replace the RoI pooling layer by a RoI align layer similar to the recent Mask R-CNN~\cite{he2017mask}. In the traditional RoI pooling layer, the region of interest coordinates are first rounded according to the stride of the convolutional features, then split into a fixed number of cells for which the coordinates are also rounded, and a max-pooling operator is applied in each cell. In contrast, the RoI align layer is designed to conserve the spatial details as it avoids these rounding operations. The features for 4 regularly sampled points per cell are obtained by bilinear interpolations and a max-pooling operator is used in each cell.
We use the same parameters as~\cite{FasterRCNN} for RPN. \new{For the classification and regression loss, with a network based on the VGG16 architecture~\cite{vgg}, 
we use 256 boxes per batch, with 32 boxes coming from 8 different
images, \ie, from more images than in the standard version.
We have more labels and, consequently, we need more diversity inside each batch.
One quarter of the boxes are on humans, the remaining ones on background.
For a network with a ResNet50 backbone~\cite{HeZRS16}, we follow the standard values from~\cite{he2017mask} and use 1 image per batch, and 512 boxes per image, with also 25\% positive examples.
In both cases, the weights are initialized with ImageNet~\cite{deng2009imagenet} pretraining.}

\begin{figure}[htb]
 \centering
 \includegraphics[width=\linewidth]{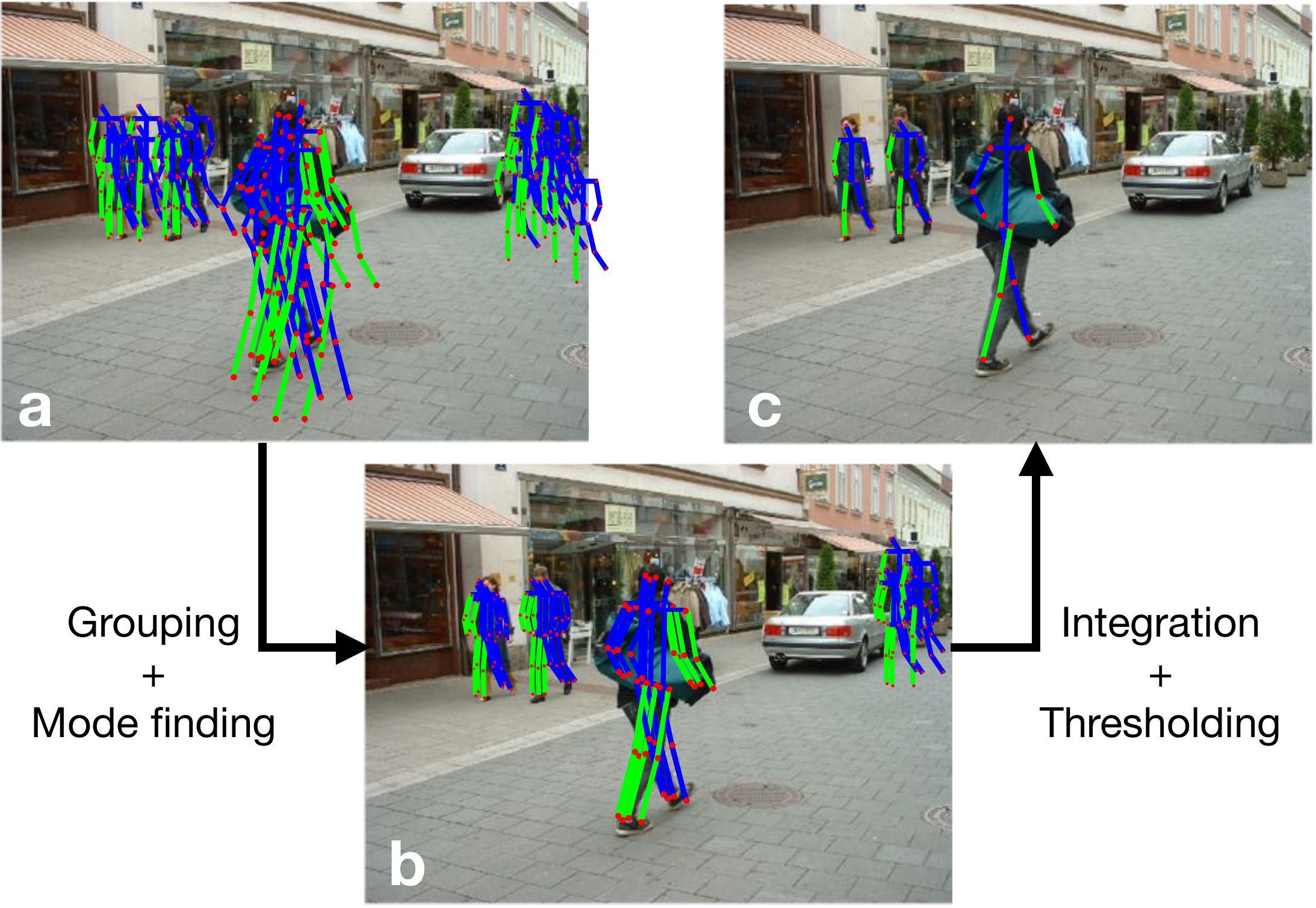}
 \caption{Illustration of the pose proposal integration (PPI). The pose proposals  (a) are grouped based on 2D overlap and 3D pose to identify the persons and the modes (b). Final pose estimates (c) are obtained by averaging the 2D poses in the selected modes and thresholding.}
 \label{fig:ppi}
\end{figure}

\subsection{Pose proposals integration}
 \label{sect:PPI}
 LCR-Net outputs a set of refined pose proposals with multiple proposals covering each person present in the image. 
One possibility is to use a non-maximum suppression
algorithm (NMS) and return the top scoring proposal for a given
region as estimated pose.  Instead, we propose to 
aggregate proposals which are close in terms of image
location and  3D pose.  We refer to this post processing stage as the pose
proposal integration (PPI), see Figure~\ref{fig:ppi}. 

\new{Each pose proposal is assigned a classification score $s(p,P)$ $=u ({c_B})$  from Equation~\ref{eqn:clsloss}. This score does not always account for the quality of the regressed pose. To penalize pose proposals  with one or several joints outside the bounding box B with respect to poses that are entirely inside the box, and consequently more likely to be accurate, we propose to rescore the proposals using}:
\begin{equation} 
\new{s'= s \; \frac{\sum_j f(p_j, B)}{J}~~,}
\label{eqn:regscore}
\end{equation}  
\new{where function $f(p_j, B)=1$ inside box $B$ and gradually decreases outside $f(p_j, B)=exp (-\mathfrak{D}^2(p_j, B)/\sigma_b^2)$,    $\mathfrak{D}(p_j, B)$ being the  distance of joint $j$ to the boundary of box $B$}. 
In practice, $\sigma_b$ is set to 25 pixels. If all the joints are inside $B$, then $s'=s$.

We start with grouping pose proposals with a sufficient spatial
overlap in the 2D image, \ie, an IoU above a certain threshold
for the bounding boxes around the 2D joints. 
We take the top scoring proposal in the image and 
determine all the pose proposals that overlap sufficiently with this top scoring
proposal. We repeat this step with the remaining pose proposals and their
top scoring elements until no pose proposals are left. The resulting groups are coherent in terms of spatial overlap but can
consist of very different 3D poses and hence the modes in 3D pose
space need to be identified. 
Let $\mathcal{P} = \{ (p,P) \}$ be the set of pose proposals in a group, each one with a classification score $s'(p,P)$.
We first pick the proposal with the highest score, 
\ie,  $(p^\ast,P^\ast) = \argmax_{(p,P) \in   \mathcal{P}} s'(p,P)$ .
We then select the set $\mathcal{Q}$ of pose proposals in the group $\mathcal{P}$,
for which the 3D distance $D_{3D}$ from  $P^\ast$ is below a threshold $T_{3D}$:
\begin{equation}
 \mathcal{Q} = \big\{ (p,P) \in \mathcal{P} ~\vert~ D_{3D}( P^\ast, P) < T_{3D} \big\} ~~.
\end{equation}
This selection ensures that we do not average poses that
belong to different modes. The PPI is thus parameterized by 2D and 3D thresholds, \ie, IoU and  $T_{3D}$ respectively.

We then obtain our final 2D pose $p$ (and similarly the 3D pose)
by averaging the 2D poses in mode $\mathcal{Q}$  weighted by their scores:
\begin{equation}
 p = \frac{1}{S} \sum_{(q,Q) \in \mathcal{Q}} s'(q,Q) \times q ~~,
\end{equation}
with $S = \sum_{(q,Q) \in \mathcal{Q}} s'(q,Q)$, sum of the individual scores.
The score for this pose $p$ is set to $S$, which results in a higher
score for poses with multiple pose proposals. 
We iterate this process, starting from the highest scored pose among
the ones that have not yet been covered by a mode. \new{The PPI stage can favor highly populated modes, which may have missed the best scored proposal, over other modes which keep the best proposals but have few other proposals to average with. Our intuition is that it is better to favor multiple coherent hypothesis than isolated top scoring ones.}
\begin{figure*}[ht]
 \centering
 \includegraphics[width=\linewidth]{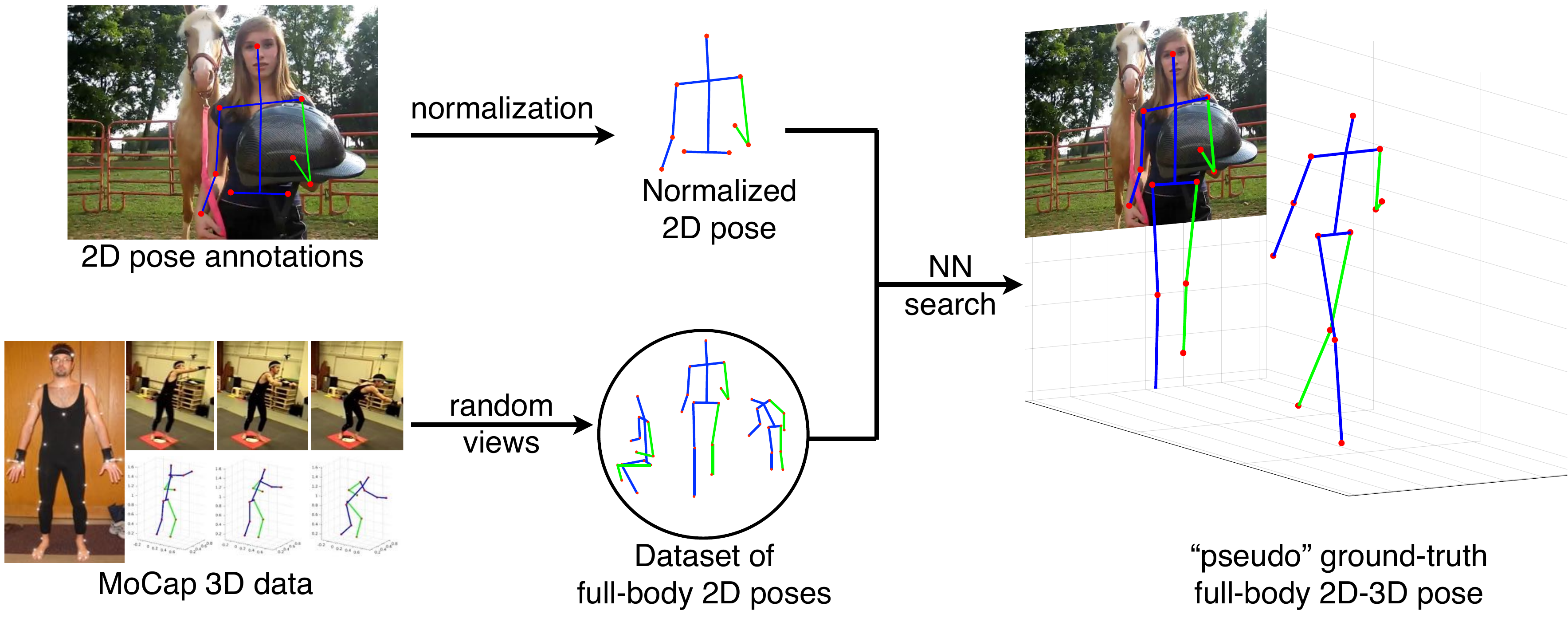}
 \caption{Pseudo ground-truth full-body 2D-3D pose annotation. From left to right: given an image with a manual 2D annotations, the pose is first normalized, then it is compared against a dataset of full-body 2D poses. These 2D poses are obtained by projecting a large corpus of MoCap 3D poses on multiple random views and normalizing them with respect to the annotated joints only. The closest pose is recovered and used (a) to define a ``pseudo'' ground-truth full-body 3D pose and (b) to complete missing annotations of the 2D pose.}
 \label{fig:NNmocap}
 \vspace{2mm}
\end{figure*}

\subsection{Pseudo ground-truth  3D pose}
\label{sub:pseudogt}

To train our network, we need full-body 2D and 3D ground-truth poses associated with each training image. 
Existing datasets with images captured in-the-wild only provide 2D joint locations of the visible joints.
Inspired by Iqbal \etal~\cite{IqbalGG16} who use 2D poses to retrieve the normalized nearest 3D poses from a motion capture dataset, we propose to infer ground-truth 3D poses from 2D annotations using a nearest neighbor (NN) search performed on the annotated joints. A similar method was recently followed in ~\cite{ChenR17} to estimate 3D pose from 2D joints locations.

A large corpus of MoCap 3D poses is first projected orthographically on multiple random virtual views to generate a very large set of 2D poses and associated orientated 3D poses $\mathcal{M} = \{ (p_m,P_m) \}_m$. Next, given an annotated 2D pose $p$, a search is performed with the normalized pose $\bar{p}=p/||p||$ to estimate the closest match, 
\ie, the 3D pose and camera view within $\mathcal{M}$ for which the 2D distance $D_{2D}$ is smallest:
\begin{equation}
(p_m^\ast,P_m^\ast) = \argmin_{(p_m,P_m) \in   \mathcal{M}} D_{2D}(\bar{p},\bar{p}_m)~~.
\end{equation}
The 3D pose of the closest match $P_m^\ast$ is then considered as ``pseudo'' ground-truth of the query 2D pose $p$. In practice, when humans are truncated or partially occluded, some joints of the 2D pose can be missing. In such cases, the normalized poses $\bar{p}$ and $\bar{p}_m$ are computed using the annotated joints only. The recovered 2D pose $p_m^\ast$ is then employed to complete missing 2D annotations so that each training instance is associated with full-body 2D and 3D annotations. See example in Figure~\ref{fig:NNmocap}.

\section{Experimental results}
\label{sec:xp}

In this paper, we address joint 2D and 3D human pose detection in natural images.
To evaluate our method, we  perform separate experiments on (a) 3D pose estimation in a controlled environment, \ie, on the Human3.6M dataset~\cite{IonescuPOS14} (Section~\ref{sub:h36}), (b) 2D and 3D pose estimation in natural images on the MPII human pose dataset~\cite{andriluka14cvpr} (Section~\ref{sub:mpii}) \new{and 3D pose estimation in the wild, \ie, on the MuPoTS dataset~\cite{MehtaSMXSPT18} (Section~\ref{sub:mupots})}.

\subsection{3D pose detection on Human3.6M}
\label{sub:h36}

\noindent{ \bf Dataset and evaluation protocols.}
 The Human3.6M dataset~\cite{IonescuPOS14} contains 3.6M human poses from 11 actors performing 17 different scripted actions. 
 The videos are captured in a controlled environment from 4 different camera viewpoints while accurate 3D poses are measured using a MoCap system. 
 Accurate 2D poses are also available for each camera view. 
 To exhaustively compare our results with the state of the art, we use the three different protocols used in the literature.  
 The first one, denoted as P1, is introduced in~\cite{KostrikovG14} and employed in \cite{RogezS16,IqbalGG16}:
 six subjects (S1, S5, S6, S7, S8 and S9) are used for training and every $64^{th}$ frame of subject S11/camera 2, \ie, a total of 928 frames, are used for testing.  
 We report the 3D pose error~(mm), averaged over the 13 joints.  Since most methods report results using more than 13 joints, we also present results for a model trained to estimate 17 joints instead of 13, adding pelvis, back, torso and neck keypoints. 
 As in~\cite{IqbalGG16}, we report a 3D pose error that measures
 accuracy of pose aligned with a rigid transformation (Align.), but also report the absolute error (Abs.). 
 The second protocol, denoted as P2, is used in~\cite{LiZC15,TekinRLF16,ZhouZLDD16}. All the frames from subjects S9 and S11 are used for testing and only S1, S5, S6, S7 and S8 are used for training.
We evaluate only on every $5^{th}$ frame as in~\cite{ZhouZLDD16}, \ie, on a test set of 110k images, as we did not observe a significant impact on performance when evaluating on all the frames. The last protocol P3, introduced by Bogo \etal~\cite{BogoKLGRB2016}, uses the same subjects for training and testing as P2. However, evaluation is performed only on sequences from camera 3 / trial 1 after rigid alignment.
\begin{figure}
 \centering 
   \begin{tabular}{cc}
 \hspace{-3mm}\includegraphics[height=0.42\linewidth]{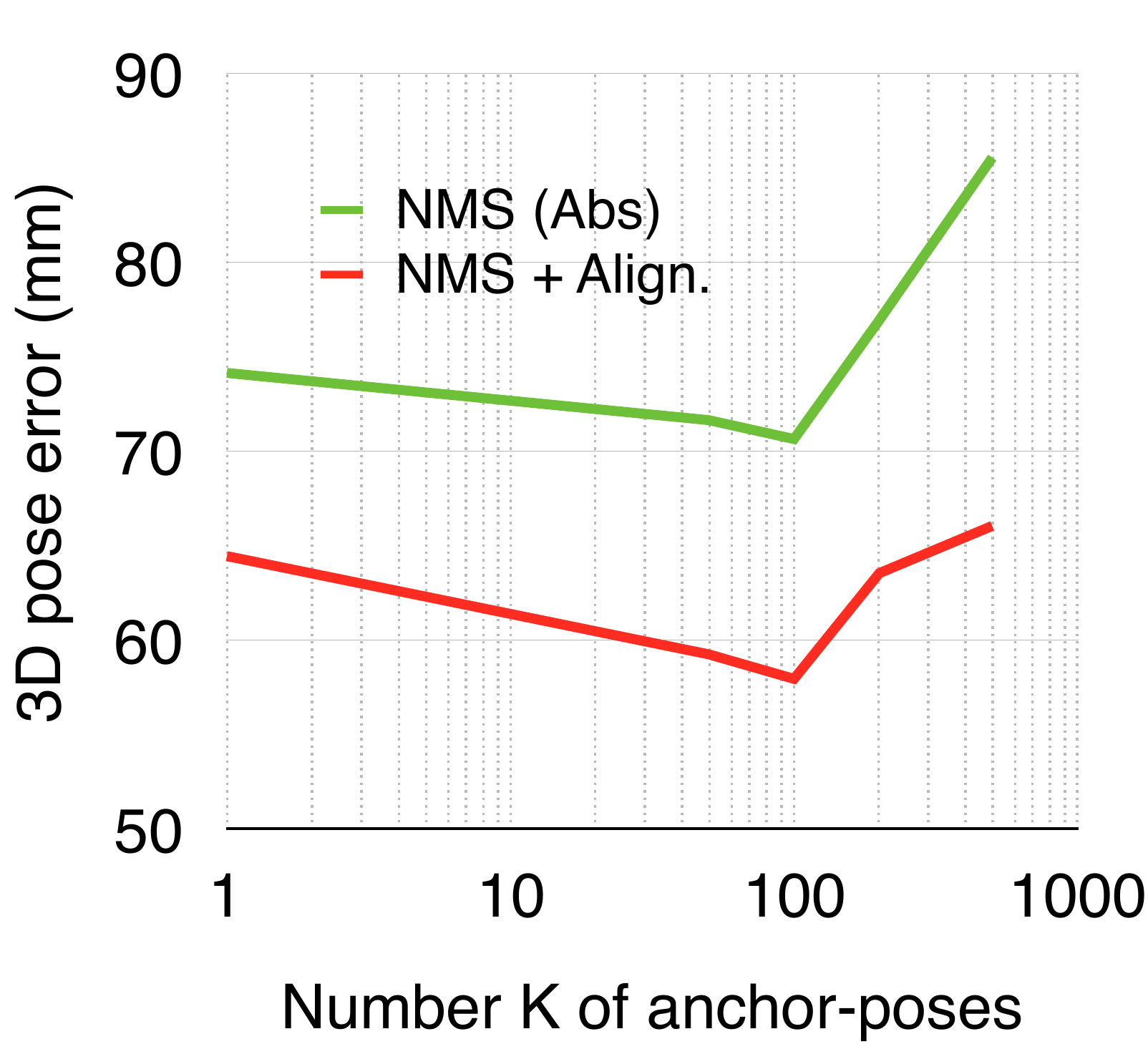}&
  \hspace{-3mm}  \includegraphics[height=0.42\linewidth]{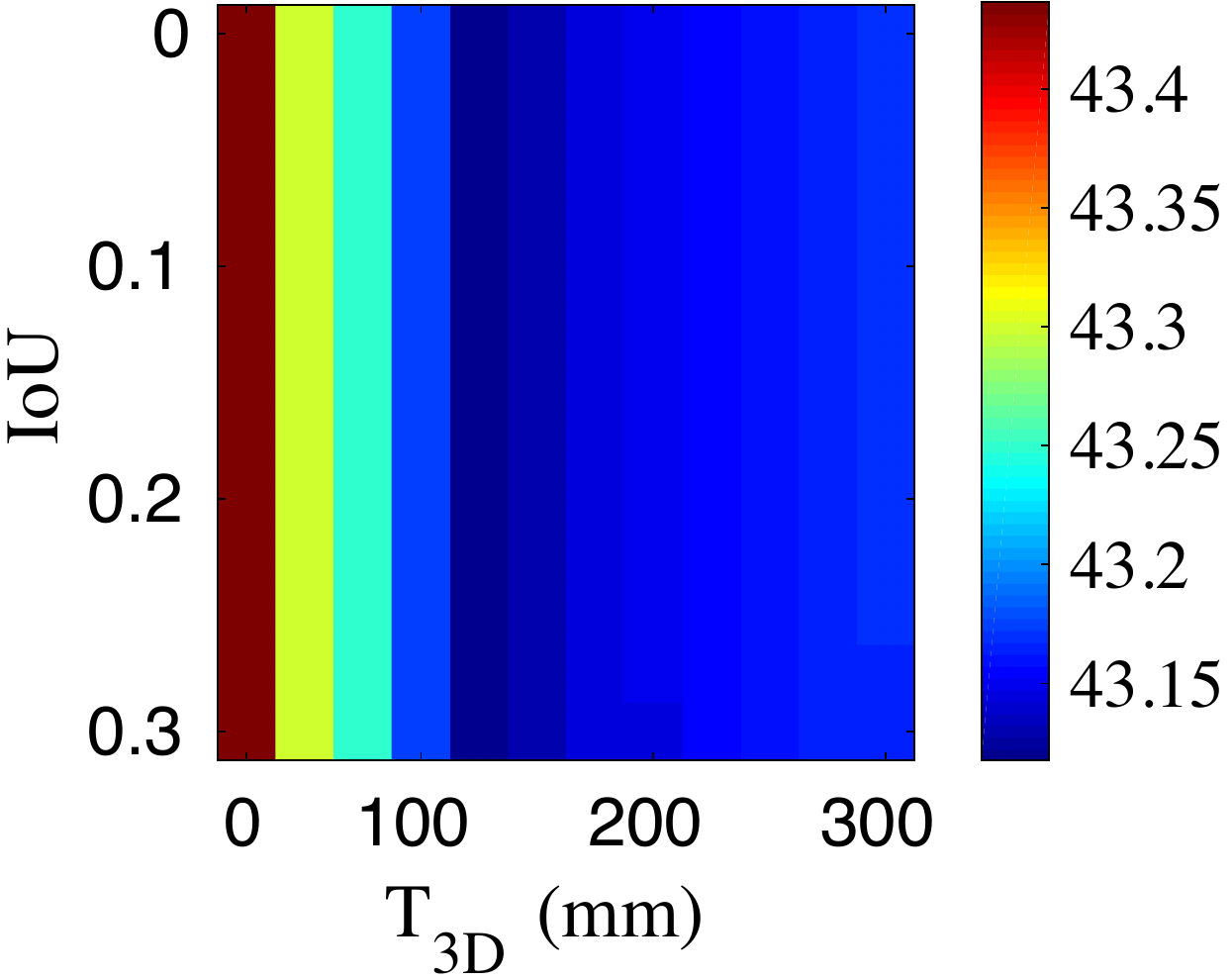}\\
  (a)&(b)
  \end{tabular}
 \caption{Average 3D pose error in mm on Human3.6M protocol P1 with respect to the number K of
anchor-poses (a) and the 2 PPI thresholds (b). Note that results on in (a) are reported  for
  NMS with/without rigid alignment for a model \new{with a VGG backbone} regressing 13 joints and trained during 100k iterations. Results in (b) are obtained after  rigid alignment with our best architecture trained to regress 17 joints.}
 \label{fig:plotH36Mk}
\end{figure}
\begin{figure*}
 \centering
   \begin{tabular}{cccc} 
    \multicolumn{4}{c}{\hspace{-2mm}\includegraphics[width=\linewidth]{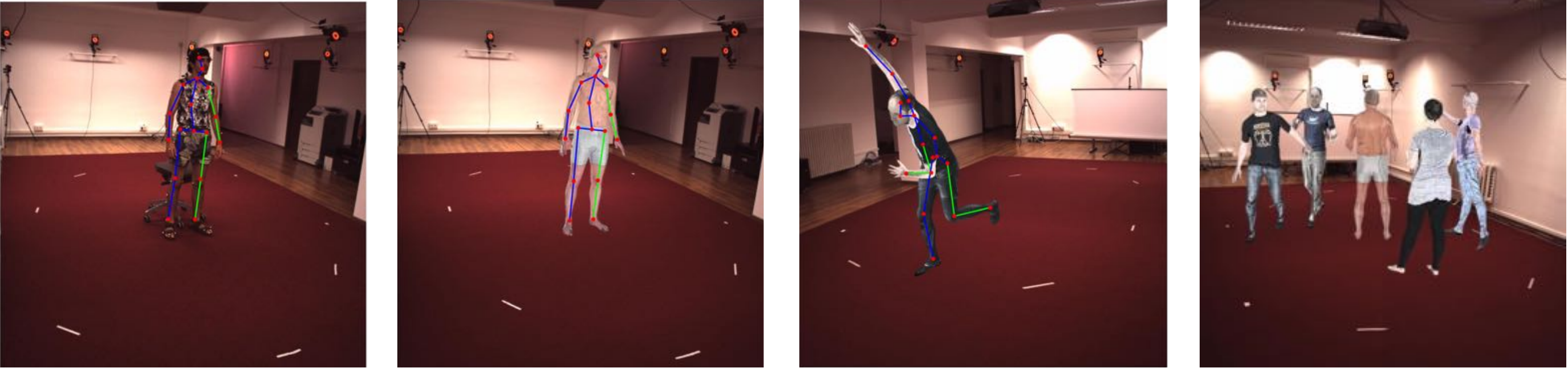}}\\   
 \hspace{15mm} (a)& \hspace{40mm} (b)& \hspace{37mm}(c)&\hspace{20 mm}(d)
  \end{tabular}
 \caption{Human3.6M real and synthetic training data. We show a training image from protocol 2 with the overplayed 2D pose in (a). In (b), we show a synthetic ``surreal''~\cite{VarolRMMBLS17} image, \ie an image obtained after rendering the SMPL model~\cite{LoperM0PB15} using the Human3.6M 3D pose from (a) and a randomly picked body shape and texture map from~\cite{VarolRMMBLS17}. Note that for more realism, the surreal image is rendered at the exact same 3D location in the MoCap room, using the camera parameters and background from the real image in (a). In (c), we show an example of image synthesized using  a 3D pose from the CMU motion capture dataset~\cite{CMUposedataset}. In (d), we show a multi-person image generated using  5 poses from the CMU MoCap dataset.}
 \label{im:synth} 
\end{figure*}
 
\noindent {\bf Anchor-poses.} 
We select a subset of the training set, \ie,
190k images and the corresponding 3D poses as in~\cite{RogezS16}, to build a set of anchor-poses
by clustering the 3D poses using $K$-means.  
Figure~\ref{fig:plotH36Mk}a shows the performance obtained when varying the number $K$ of
anchor-poses   
with a simple NMS, 
\ie, taking the top scoring pose proposal as 3D pose estimate.
Best performance is obtained for $K$=100. 
When $K$ is too small, for instance if $K$=1 which corresponds to a standard regression,
the number of anchor-poses might not be sufficient to cover the pose
space. When $K$ becomes too large, the error also increases since the
anchor-poses are too similar, resulting in ambiguities in the classification. We select $K$=100 classes for the remaining experiments on Human3.6M.

\noindent {\bf Additional synthetic training data.} One of the conclusions in the earlier version of this work~\cite{RogezWS17} was that LCR-Net required a significant amount of training data that could be generated through synthesis. To augment the training set with synthetic images  with associated 3D poses, we render the SMPL 3D human mesh model~\cite{LoperM0PB15}. For more realism, we render these images in the Human3.6M capture room using background images and camera parameters provided with the data (see Figure~\ref{im:synth}). We generate images for a same quantity of poses and consider two sets of 3D MoCap poses: a) the same poses from Human3.6M to add appearance variations (Figure~\ref{im:synth}b) and b) poses from the CMU dataset~\cite{CMUposedataset} to add variations both in terms of appearance and poses (Figure~\ref{im:synth}c-d). To ensure a balanced training set, we sample CMU poses in areas  of the pose space that are less populated by Human3.6M poses. In both cases, we use the SMPL body parameters and texture maps from~\cite{VarolRMMBLS17}.  The SMPL kinematic model is somehow different from the Human3.6M 3D model: some of the 17 joints from Human3.6M poses do not correspond exactly to their SMPL counterparts (e.g., head, hips and shoulders) while others are simply missing (neck and torso). To tackle this issue, we trained a regressor from SMPL to Human3.6M poses using the body parameters estimated by~\cite{VarolRMMBLS17}, to ``correct'' the misplaced or missing joints for the CMU-based images for which we do not have Human3.6M-like pose annotations. We obtained satisfactory 17-joint poses for all the synthesized images. See examples in Figure~\ref{im:synth}c-d. In total, we obtained a training set of 557k images. We trained   VGG-based models for 500k iterations \new{(roughly 8 epochs as there 8 images per iteration) using SGD,  300k  iterations at a learning rate lr=$10^{-3}$ and 200k with lr=$10^{-4}$. The models with ResNet50 were trained for 2.7 million iterations with one image at each iteration, 1.8 million at lr=$10^{-3}$, then 0.6 million iterations at lr=$10^{-4}$ and 0.3 million iterations at lr=$10^{-5}$.}

\begin{figure*}
 \centering
 \includegraphics[width=\linewidth]{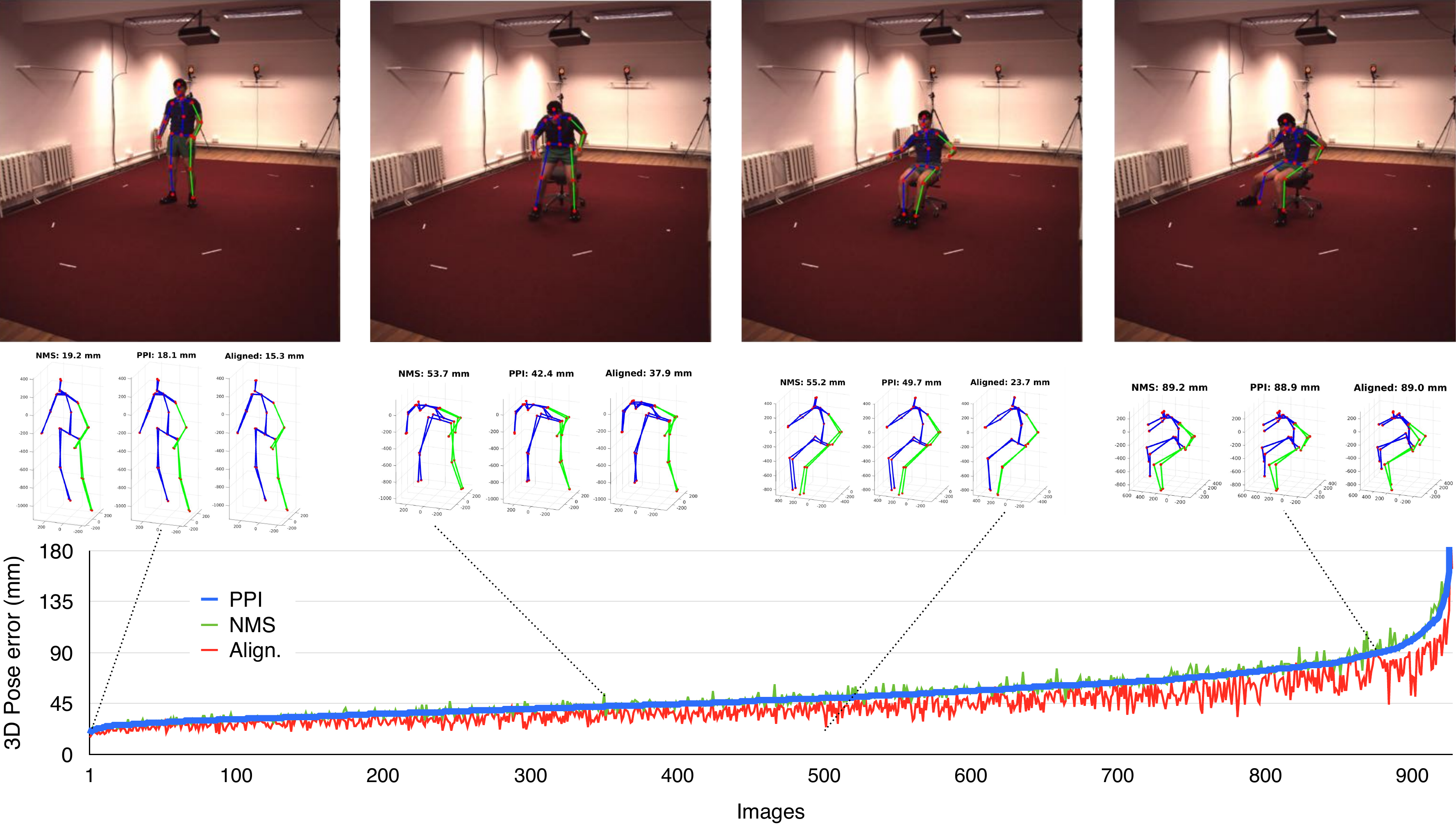}
 \caption{Average 3D pose error  on Human3.6M  test images (protocol
   P1). We order the examples by increasing error of PPI results (blue) and also report the performance with a simple NMS (green) and after rigid alignment of the PPI estimation (red). 
 We show qualitative results for 4 particular cases, from left to right: 
(a) an image where NMS estimation is already accurate, thus PPI and alignment do not further improve, 
(b) a case in which the PPI achieves an accurate pose estimate, 
(c) a case where PPI does not improve over NMS but the alignment helps to correct the pose estimate and 
(d) a failure case where the pose is not satisfactory, even after rigid alignment. 
 For each case, we show the image with the estimated 2D pose (with PPI). We also show the 3D poses estimated by NMS, PPI and after alignment overlaid with the ground-truth 3D pose. 
 }
 \label{fig:resH36Mk}
\end{figure*}

\noindent {\bf Impact of PPI.} We merge poses that are (a) highly overlapping in 2D, \ie for which the bounding boxes intersection over union is over the IoU threshold and (b) close in 3D pose space, \ie whose 3D Euclidean distance is below $T_{3D}$. 
We  experimentally set $T_{3D}$ to $125$ mm and found that the IoU
threshold has no influence on the performance for this dataset (see Figure~\ref{fig:plotH36Mk}b), as only
one individual is observed and all highly scored proposals are localized on the subject. 
In most cases, the highest scoring pose proposal (NMS) is already an accurate
 estimation but, on average, the improvement achieved by our
PPI over the NMS estimates is non negligible. On protocol P1, we obtain an average error of  $54.2$ mm after NMS and $53.5$ mm after PPI ($43.7$ mm and $43.1$ mm after rigid alignment) when evaluating on 17 joints. 
In Figure~\ref{fig:resH36Mk}, we show some qualitative results where
examples are sorted by increasing 3D pose error.
A green upward peak with respect to the blue curve corresponding to
PPI indicates an important improvement by the PPI,
whereas a red peak
downward indicates poses where the rigid alignment helps correct the most. 
For the 928 test frames of protocol P1, less than 20 have an error equal to or greater than 90 mm. 
This occurs in cases of unseen poses in the training set, see
rightmost example in Figure~\ref{fig:resH36Mk}.

\begin{figure}[tb]    
       \centering  
    \begin{tabular}{cc}
     \hspace{-3mm}\includegraphics[width=0.5\columnwidth]{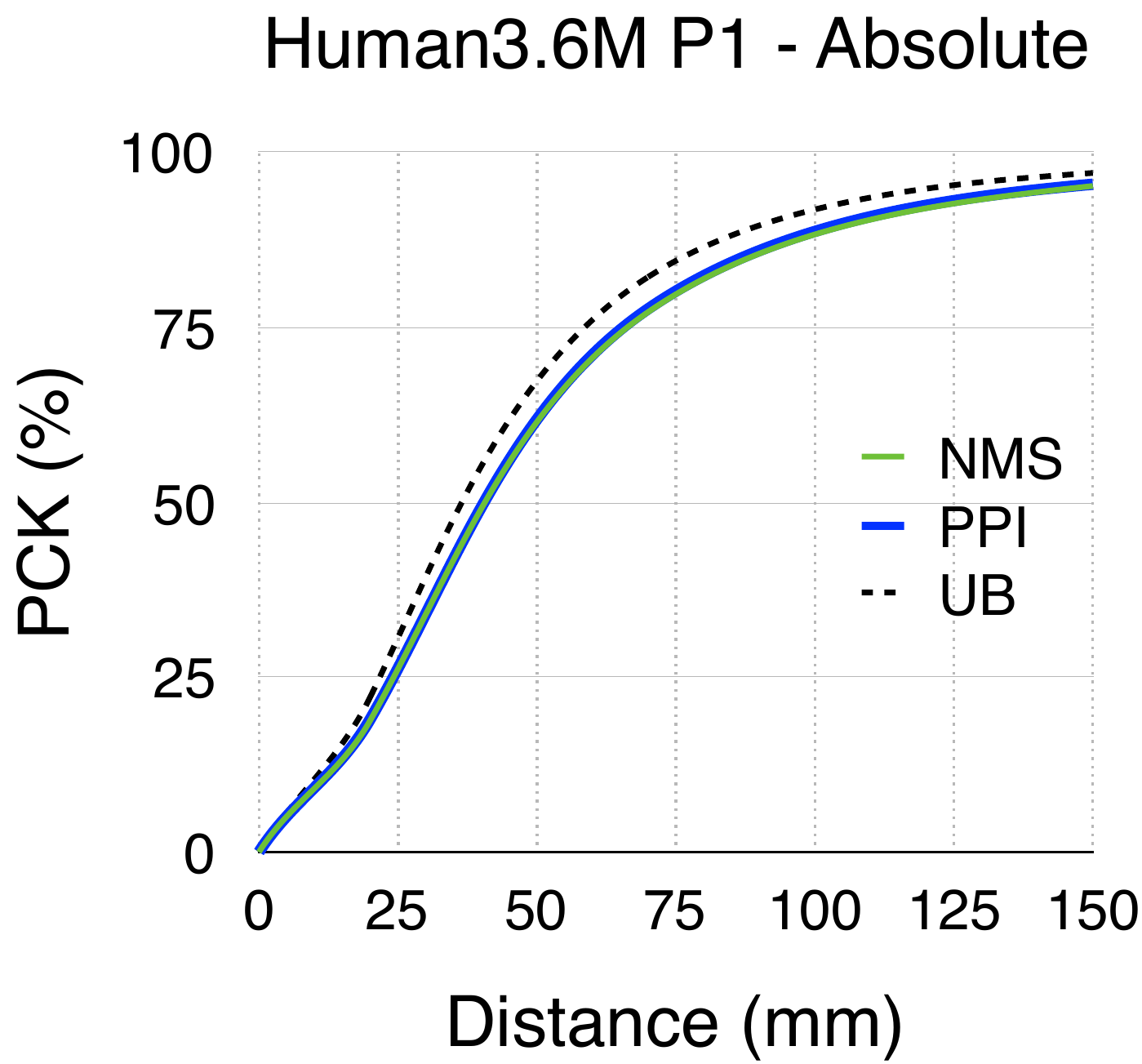}  &
  \hspace{-3mm}\includegraphics[width=0.5\columnwidth]{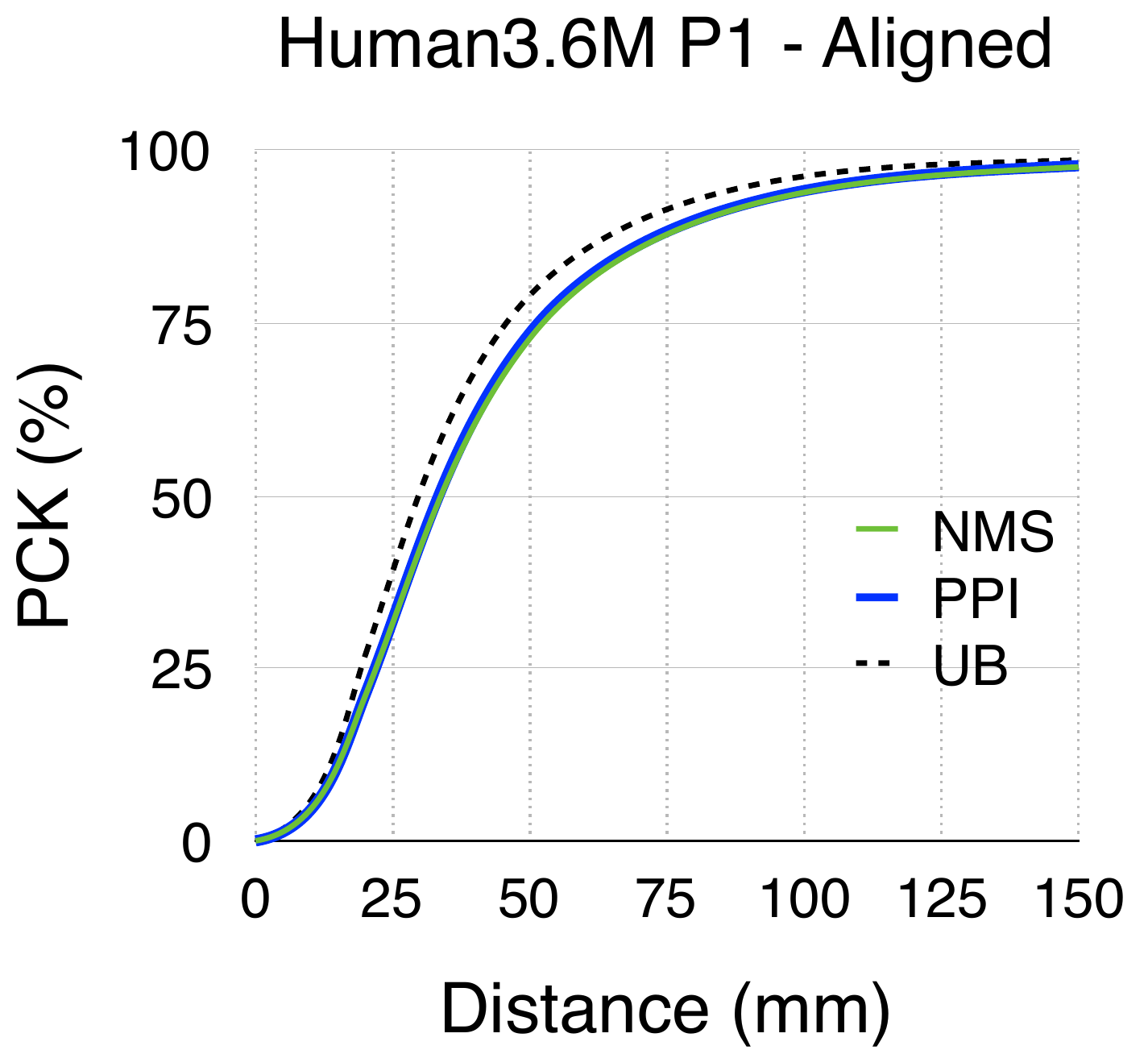}  \\
  (a) & (b) 
  \end{tabular}
\caption{Average Percentage of Correct Keypoints PCK (\%) on Human3.6M protocol P1.  Detection rate with respect to the  distance to ground truth 3D joints is given for PPI, NMS and the Upper bound (UB), \ie, taking the pose proposal closest to ground-truth pose. Performances are given before (a) and after (b) rigid alignment to the ground-truth poses.}
     \label{fig:plotsPCKH36M} 
 \end{figure}

\noindent {\bf Ablative analysis.} Our complete method, called LCR-Net++, significantly improves over the initial and simpler version of LCR-Net (as published in~\cite{RogezWS17}), \ie, a model trained without synthetic data and an architecture \new{with a VGG16 backbone} that does not include iterative refinement, RoI alignment and rescoring of the pose proposals. To better understand the origin of this improvement, an ablative analysis is provided in Table~\ref{tab:H36M_ablative} for a model trained to estimate 13 joints.  We can see that the biggest improvements are obtained when adding synthetic images to the training set. By adding variability in terms of appearance, \ie, adding synthetic data rendered using Human3.6M poses, we decrease the 3D error by $15$ mm. We can see that the gap between Abs. and Align. results is smaller ($10.9$ mm vs $16.1$ mm without using synthetic data), meaning that we better estimate the camera viewpoint. Adding synthetic training images rendered from new poses (CMU) further improves the performance by another $5$ mm. This validates the fact that our approach requires a large and varied training set in terms of pose and appearance. The RoI alignment and iterative process do not help improve the performance on Human3.6M significantly as the 3D estimations are already quite accurate. Rescoring the pose proposals following Equation~\ref{eqn:regscore} (in Section~\ref{sect:PPI}) allows the NMS to select better pose proposals and the PPI to produce better pose estimates after integration, \ie, the 3D error decreases by $3$ mm. \new{Using a ResNet50 backbone instead of VGG16, does not help improve the performance substantially. A VGG16 network has sufficient learning capabilities for  a controlled environment.} In Figure~\ref{fig:plotsPCKH36M}, we report the Percentage of Correct Keypoints (PCK), \ie, the ratio of joints for which the error is below a threshold, on Human3.6M protocol P1. When computing the upper bound, \ie, taking the pose proposal closest to ground-truth pose and thus simulating a perfect scoring, we observe a boost in performance, both before and after rigid alignment. This indicates that even after applying our rescoring function, the top scoring pose proposals are not always the ones that best explain the input images. In some cases, information from previous frames could help disambiguate and adequately rescore the pose proposals. In future work, our method could  be extended to leverage such additional temporal information, which should further improve the performance. 
\begin{table}[tb]
\centering
\resizebox{\linewidth}{!}{
\begin{tabular}{lccc}
\toprule
    &   NMS   &  PPI &     PPI   \\
      &  (Abs.)   &  (Abs.)&    (Align.)   \\
\midrule
LCR-Net \new{ (with VGG16 backbone) } ~\cite{RogezWS17}    & 89.8& 87.7 &71.6 \\ 
 \hspace{0.5cm} + synth Human3.6M    &  73.3 & 73.9 & 63.2 \\
\hspace{0.75cm}  + synth CMU    &  68.5 & 68.9 & 59.3 \\
\hspace{1cm}  + RoI align    &  68.3 & 69.3 & 59.6 \\
\hspace{1.25cm}  + iterative estimation    &  67.7 & 68.7  & 59.3 \\
\hspace{1.5cm}  + rescoring  \new{ (\textbf{LCR-Net+}) }     &  66.8 &  65.8  &56.4 \\
\hspace{1.25cm}\new{+ ResNet50 backbone }   &  \new{ 68.2} & \new{ 65.4}  & \new{ 54.4 }\\
\hspace{1.5cm}  \new{+ rescoring  (\textbf{LCR-Net++}) } & \new{ \bf{67.2}}& \new{ \bf{65.4} } &\new{ \bf{54.3} } \\
\bottomrule
\end{tabular}
}
\vspace{0mm}
\caption{Ablative analysis on Human3.6M protocol P2 (evaluating on 13 joints).  We evaluate the performance of LCR-Net when adding the different modifications introduced in this work compared to the simpler version published in~\cite{RogezWS17}, \ie,  with a RoI pooling layer and trained on Human3.6M training set only. 
For each tested model/architecture, the average absolute  3D pose error (mm)  is reported for NMS, and also  PPI before (Abs.) and after
rigid 3D alignment (Align.) }
\label{tab:H36M_ablative}
\end{table}
\begin{table}[tb]
\centering
\resizebox{\linewidth}{!}{
\begin{tabular}{lccccc}
\toprule
 \multirow{2}{*}{ Methods (num. joints) }&  P1  & {\bf P1} &  {\bf P2}   &  P2 &   {\bf P3  } \\
    &   (Abs.)   &  {\bf (Align.)}  &   {\bf (Abs.)}   &  (Align.)   &  {\bf (Align.)}  \\
\midrule
Bo \& Sminchisescu~\cite{BoS10}  (14 jts)   & - & 117.9 &- & - & -\\
Kostrikov \& Gall~\cite{KostrikovG14}  (14 jts)   &  - &115.7 &- & -  & - \\
Iqbal~\etal.~\cite{IqbalGG16}   (14 jts)  & - & 108.3 &- & -  & - \\
Du~\etal \cite{DWLHGWKG16} (14 jts) &   - & - & 126.5 & -& - \\
Bogo~\etal~\cite{BogoKLGRB2016}   (14 jts)   & - & -  & - & -  & 82.3\\
Rogez \& Schmid~\cite{RogezS16}  (13 jts)  &126 & 88.1 & 121.2 & 87.3&-\\ 
Chen \& Ramanan~\cite{ChenR17}  (14 jts)   & - & 82.7  & 114.2& -  & -\\
Rogez \& Schmid~\cite{RogezS18}  (13 jts)  &116.7 & 90.1 & 110.6 & -&-\\ 
Nie~\etal~\cite{NieWZ17} (13 jts)    & - & 79.5  & 97.5 & -  &-\\ 
Moreno-Noguer~\cite{Moreno17} (14 jts)      & - & 74.0  & 85.6 & -  & 81.5\\
\midrule
LCR-Net~\cite{RogezWS17}  (13 jts)     & 63.2&53.4& 87.7 &71.6&72.7\\
\bf{LCR-Net+} (13 jts)    & 56.8  &  48.3  & 65.8  & 56.4  & 57.2 \\
\new{\bf{LCR-Net++} (13 jts)  }  &\bf{54.6} & \bf{45.8} &\bf{65.4} &\bf{54.3} &\bf{56.5} \\
\toprule
Li~\etal~\cite{LiC14} (17 jts)   & - & - &  136.5 & -& - \\
Li~\etal~\cite{LiZC15} (17 jts)   & - &  - & 122& -& - \\
Tekin \etal~\cite{TekinRLF16} (17 jts)   & - & - &  125.0 & -& - \\
Park~\etal \cite{ParkHK16} (17 jts) &   - & - & 117.3 & -& - \\
Zhou \etal~\cite{ZhouZLDD16} (17 jts) &  - & - &  113.0 & -& - \\
Zhou~\etal~\cite{ZhouSZLW16} (17 jts)   & - & -  &107.26 & -  & -\\
Sanzari~\etal \cite{SanzariNP16}  (17 jts) & - & - &93.1 & - & -\\
Tome~\etal~\cite{TomeRA17}   (17 jts)   & - &70.7  & 88.4 & -  &79.6\\ 
Mehta~\etal~\cite{MehtaRCFSXT17}   (17 jts)  & 72.8 & -  & 74.14 &-&-\\  
Pavlakos~\etal~\cite{PavlakosZDD17}  (17 jts)   & - & -  &71.9 & 51.9  & -\\
Tekin~\etal~\cite{TekingMSF17} (17 jts)    & - & -  & 70.81 & 50.1  &-\\ 
Katircioglu~\etal~\cite{KatirciogluTSLF18} (17 jts) & - & -  &65.4 & -  &-\\ 
Zhou~\etal~\cite{ZhouHSXW17}  (16 jts)   & - & -  & 64.9& -  &-\\ 
Martinez~\etal~\cite{MartinezHRL17} (17 jts)    & - & -  & 62.9& \bf{47.7}  &-\\ 
Sun~\etal~\cite{SunSLW17} (16 jts)    & - & 48.3  & - & -  &-\\ 
Kinauer~\etal~\cite{KinauerGCK17} (16 jts) &  - & 45.9  & - & 54.5  &-\\ 
\midrule
\bf{LCR-Net+}   (17 jts)   & \bf{53.5} & 43.1   & \bf{61.2}  &49.4&\bf{50.5}\\
\new{\bf{LCR-Net++}  (17 jts)  }  & 53.9 & \bf{42.7}  & 63.5  &49.2& 51.1 \\
\bottomrule
\end{tabular}
}
\vspace{1mm}
\caption{Comparison with state-of-the-art results on Human3.6M for 3 different protocols. 
The average 3D pose error (mm) is reported before (Abs.) and after
rigid 3D alignment (Align.) for protocols P1 and P2. 
See text for details.  We group the methods  according to the number of joints that are evaluated (13-14 or 16-17). The errors are globally higher with  P2 and P3 that provide less training subjects and have a larger and more varied test set. 
}
\label{tab:H36M_sota}
\end{table}
\begin{table*}[htb]
\resizebox{\linewidth}{!}{
\begin{tabular}{lcccccccccccccccc}
\toprule
Method  (num. joints) & Dir. & Disc. & Eat & Greet & Phone  & Photo &  Pose & Purch. & Sit & SitD. & Smoke &   Wait & Walk  & WalkD. & WalkT. &  Avg. \\
\midrule
Rogez \& Schmid~\cite{RogezS16} (13 jts)&  94.5  & 110.4 & 109.3 & 143.9 & 125.9 & 160.3 & 95.5 & 89.8 & 134.2 & 179.2 & 123.8  & 133.0 & 77.4 & 129.5 & 91.3 &  121.2  \\
Chen \& Ramanan~\cite{ChenR17}  (14 jts)   & 89.9 & 97.6 & 90.0 & 107.9 & 107.3 & 139.2 & 93.6 & 136.1 & 133.1 &  240.1 & 106.7 & 106.2 &114.1 & 87.8 & 90.6&114.2\\
Rogez \& Schmid~\cite{RogezS18} (13 jts)& 87.7  & 100.7 & 93.6 & 139.6 & 107.9 & 155.2 & 88.1 & 78.9 & 119.0 & 171.9 & 107.4  & 130.7 & 71.6 & 114.6 & 83.1 &  110.6  \\
Nie~\etal~\cite{NieWZ17} (13 jts)    &  90.1 & 88.2 & 85.7 & 95.6 & 103.9 & 92.4 & 90.4 & 117.9 & 136.4 & 98.5 & 103.0 & 94.4 & 86.0 & 90.6 & 89.5 & 97.5\\ 
Moreno-Noguer~\cite{Moreno17} (14 jts)      &  67.5 & 79.0 & 76.5 & 83.1 & 97.4 & {\bf 74.6} & 72.0 & 102.4 & 116.7 & {\bf 87.7} & 100.4 & 94.6 & 75.2 & 82.7 & 74.9 & 85.6 \\
\midrule
 LCR-Net \cite{RogezWS17} (13jts)   &  76.2 &  80.2 &  75.8 & 83.3& 105.7  &  92.2 &79.0 &  71.7 &   105.9 &   127.1 &  88.0 &  83.7 & 64.9 &  86.6 & 84.0 &  87.7 \\
 \bf{LCR-Net+}  (13jts) &   {\bf 53.4} &  {\bf 59.1} &  61.8 &  59.6 &  72.3 &  78.3 &{\bf  54.1} & {\bf 55.7} &   95.6 &  99.5& {\bf 68.7}&  59.4 &  {\bf 47.1 } &  66.3 &  56.4  &  65.8 \\
 \new{\bf{LCR-Net++}  (13jts)} &   55.5 &  59.4 & {\bf 61.0} & {\bf 59.2} & {\bf 70.2} & {\bf 77.5} & 55.0 & 56.6 &  {\bf 88.4 }&  101.0& 69.0& {\bf 59.3} &  47.9 & {\bf 65.6} & {\bf 55.5} &  {\bf 65.4} \\
\toprule
Sanzari~\etal \cite{SanzariNP16}   (17 jts)   & {\bf 48.8}& 56.3 & 96.0& 84.8  & 96.5 &   105.6  & 66.3& 107.4 & 116.9 & 129.6 & 97.8 & 65.9 & 92.6 & 130.5& 102.2 & 93.1   \\ 
Tome~\etal~\cite{TomeRA17}   (17 jts)   & 65.0 & 73.5 & 76.8& 86.4 & 86.3 & 110.7 & 68.9 & 74.8 & 110.2 & 173.9 & 84.9 & 85.8 & 71.4 & 86.3 & 73.1 & 88.4   \\  
Pavlakos~\etal~\cite{PavlakosZDD17}  (17 jts)   & 67.4 & 71.9 & 66.7 & 69.1 & 71.2 & 77.0 & 65.0 & 68.3 & 83.7   & 96.5 & 71.7 & 65.8 & 59.1 & 74.9 & 63.2 & 71.9 \\
Tekin~\etal~\cite{TekingMSF17} (17 jts)              & 53.9 & 62.2 & 61.5 & 66.2 & 80.1 & 79.5 &  64.6 & 83.2 & {\bf 70.9} & 107.9 & 70.4 & 68.0 & 52.8 & 77.8 & 63.1 & 70.8 \\ 
Katircioglu~\etal~\cite{KatirciogluTSLF18} (17 jts) & 54.9 &63.3 &{\bf 57.3} & 62.3&70.3&77.4&56.7&57.1&79.0&97.1&64.3&61.9&49.8&67.1&62.3&65.4 \\ 
Zhou~\etal~\cite{ZhouHSXW17}  (16 jts)   & 54.8 & 60.7 & 58.2 & 71.4 & {\bf62.0} & {\bf65.5 }& 53.8 &55.6 & 75.2 & 111.6 & 64.15 & 66.05 & 63.2& 51.4 & 55.3 & 64.9  \\ 
Martinez~\etal~\cite{MartinezHRL17} (17 jts)    &  51.8 & 56.2 & 58.1 & 59.0 & 69.5 & 78.4 & 55.2 & 58.1 & 74.0 & 94.6 & {\bf 62.3 }& 59.1 & 49.5 & 65.1 &  {\bf  52.4} & 62.9\\ 
\midrule
\bf{LCR-Net+}  (17jts) &    50.9 &  {\bf 55.9} &  63.3 & {\bf 56.0} &  65.1 &  70.7 &{\bf  52.1} & {\bf 51.9} &  81.1 &  91.7 & 64.7 & {\bf 54.6} &  {\bf 44.7 } & {\bf 61.1} & 53.7 &  {\bf61.2} \\
\new{\bf{LCR-Net++} (17jts)} &    55.9 &  60.0 &  64.5 & 56.3 &  67.4 &  71.8 &  55.1 &  55.3 &  84.8 &  {\bf 90.7}& 67.9 &  57.5 &  47.8 &  63.3 & 54.6 &  63.5 \\
\bottomrule
\end{tabular}
}
\vspace{1mm}
\caption{Per-class results on Human3.6M  protocol P2 (without pose
  alignment). We report 3D pose error results (mm) for recently published works that provide per-class performance and employ a single ``general'' model, \ie, a single model covering the 15 actions.}
\label{tab:h36classP2}
\end{table*}

\noindent {\bf Detailed comparison with the state of the art.} 
We now extensively compare our method with the state of the art.
First, Table~\ref{tab:H36M_sota} compares our complete method (LCR-Net++) to other recent competing approaches on the three protocols P1, P2 and P3. We also compare with the simpler version of LCR-Net~\cite{RogezWS17} \new{ and its improved version LCR-Net+ (both using a VGG16 backbone)}. For a fair comparison, we group methods that consider 13-14 or 16-17 joints and do not include results where a different model was trained for each action.
First, we can observe that the average 3D pose error obtained \new{by our architectures (LCR-Net+/++)} decreases when considering more keypoints. These additional joints, \ie,  pelvis, back, torso and neck keypoints, are easier to estimate compared to extremities of limbs such as wrists and ankles. Adding them in the computation of the pose error artificially improves the performance. \new{When using a ResNet50 backbone (LCR-Net++), the performance is on-par with the VGG16 backbone (LCR-Net+) with a slight improvement with 13 joints and a slight drop with 17 joints.
Overall, we}  outperform all other methods  for the 3 protocols of the literature and significantly improve over our previous results, especially on protocol P2 ($65.4$ mm vs $87.7$ mm  with 13 joints)  
which is the most difficult one as less training subjects are available and  a larger and more varied test set is considered. On this protocol, we establish a new state-of-the-art performance both with 13 ($65.4$ mm obtained with LCR-Net++) and 17 joints ($61.2$ mm obtained with LCR-Net+) and outperform all previously published methods, including very recent work,   
despite the fact that (a) we also perform localization, in
contrast to most methods such as~\cite{RogezS16,ZhouHSXW17,PavlakosZDD17} that
assume a bounding box annotation of the human and (b) we propose an end-to-end architecture trained with Human3.6M images only while other methods rely on off-the-shelf 2D pose detectors~\cite{Moreno17,TekingMSF17,MartinezHRL17,ChenR17}.  
Note that we do not include in this table the results reported by~\cite{SunSLW17}  on P2 Abs. ($59.1$ mm)  as the authors did not follow the exact same protocol and evaluated on a much smaller subset of the test images, 9.6k randomly sampled images instead of 110k, making the comparison unfair. 
When adding a rigid transformation for protocol P2, the method of~\cite{MartinezHRL17} achieves a slightly better performance than ours, whereas LCR-Net+ performs better without alignment. This means that their estimation of the camera viewpoint is less accurate than ours and that aligning the poses in 3D helps to correct this lack of accuracy.
We   present a per-class comparison on  protocol P2 in Table~\ref{tab:h36classP2}.
Compared to methods estimating 13-14 joints, LCR-Net++ is state of the art for 13 out of 15 actions and only performs lower than~\cite{Moreno17}  for ``taking photo'' and ``sit down'' actions.  In the case of 16-17 joints, our method is state of the art for 8 out of 15 actions.   
The methods from~\cite{Moreno17,TekingMSF17,ZhouHSXW17,MartinezHRL17} or~\cite{KatirciogluTSLF18} report better performance for the remaining 7 actions. Katircioglu \etal~\cite{KatirciogluTSLF18}  leverage temporal information. All the other methods rely on  heatmaps or 2D joints  detected by~\cite{WeiRKS16} or ~\cite{NewellYD16} while our architecture is trained end-to-end using only the Human3.6M training set  and synthetic data.

\subsection{2D and 3D pose detection on MPII}
\label{sub:mpii}

\noindent{ \bf Datasets and evaluation protocols.}
We now present experimental results for 2D and 3D pose detection in
real-world images.
We use the challenging MPII human pose dataset~\cite{andriluka14cvpr} that consists of around 40k annotated 2D poses in around 25k images (17,4k for training and 7k for testing).
It contains a large variety of camera viewpoints and poses,
originating from around 400 different actions.
Each scene can contain multiple people, that are often occluded or truncated by the image boundary.
This makes the dataset challenging for human pose estimation.
While most other papers on 3D pose estimation
only show qualitative examples on real images, we analyze our results
on a validation set of 1k images that we used for both
single (1,088 poses) and multi-person (209 groups) protocols. This set
is obtained by randomly splitting the training dataset to create a
training set of 16,4k images and a validation set of 1k images, making
sure that images from the same video all belong to the same set. 
For training, we also use the annotated images from LSPE as in~\cite{PishchulinITAAG15, RogezS16},  a subset of 17k images from Human3.6M as in~\cite{RogezS16} and the training set of the MS Coco dataset~\cite{LinMBHPRDZ14}. After mirroring, we obtain a training set of 161k images with around 290k annotated humans. 
To understand the influence of the training data on the performance,  we further increased the size of the training set by adding synthetic images rendered using the CMU Mocap dataset as in Section~\ref{sub:h36}. This time, we synthesized 40k multi-person images with 186k humans (an average of 4-5 persons per image) as the example depicted in Figure~\ref{im:synth}d. 
We  trained the VGG-based models for 500k iterations ($\approx$20 epochs),  300k  iterations at a learning rate lr=$10^{-3}$ and 200k with lr=$10^{-4}$. \new{ The models with a ResNet50 backbone were trained for 2.7 million iterations.}
For single-person pose estimation, we report the results using the  PCKh metric that measures the ratio of estimated joints for which the distance to the ground-truth is below a threshold.
The standard threshold is set to half the size of the head, \ie, PCKh@0.5. In this setting, most methods use person localization information before computing the pose.
In our case, we detect the poses for the entire image and use the localization information only for evaluation, \ie, to select the pose that corresponds to each ground-truth.
For multi-person evaluation, we follow the standard protocol and evaluate the average precision (AP). Both PCKh and AP are averaged over 14 joints, the $14^{th}$ joint (top of the head) being extrapolated from our 13-joint pose.
\begin{figure}[h] 
   \centering   
  \begin{tabular}{cc} 
  \hspace{-0.4cm} \includegraphics[width=0.24\textwidth]{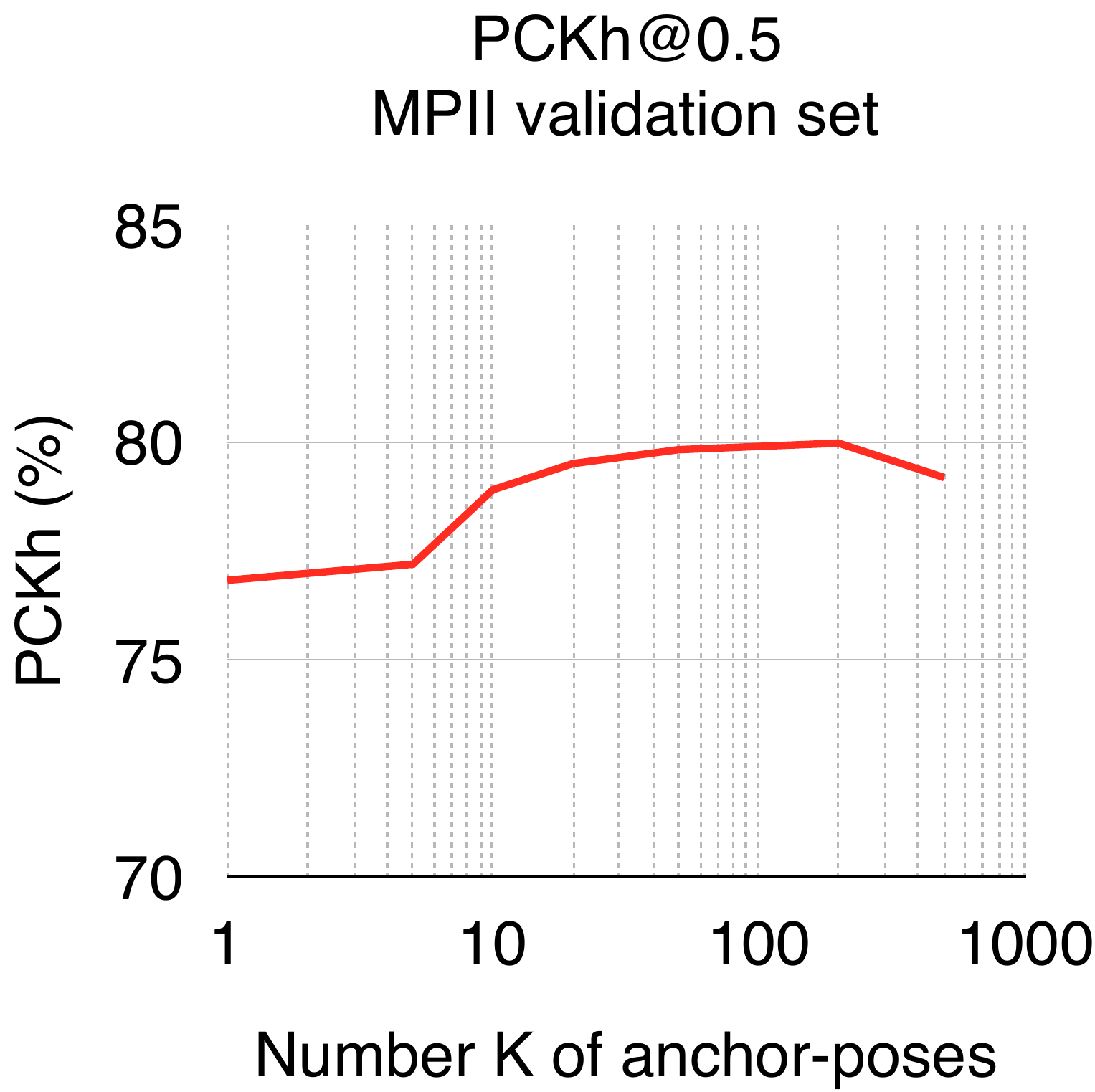}&
  \hspace{-0.4cm} \includegraphics[width=0.24\textwidth]{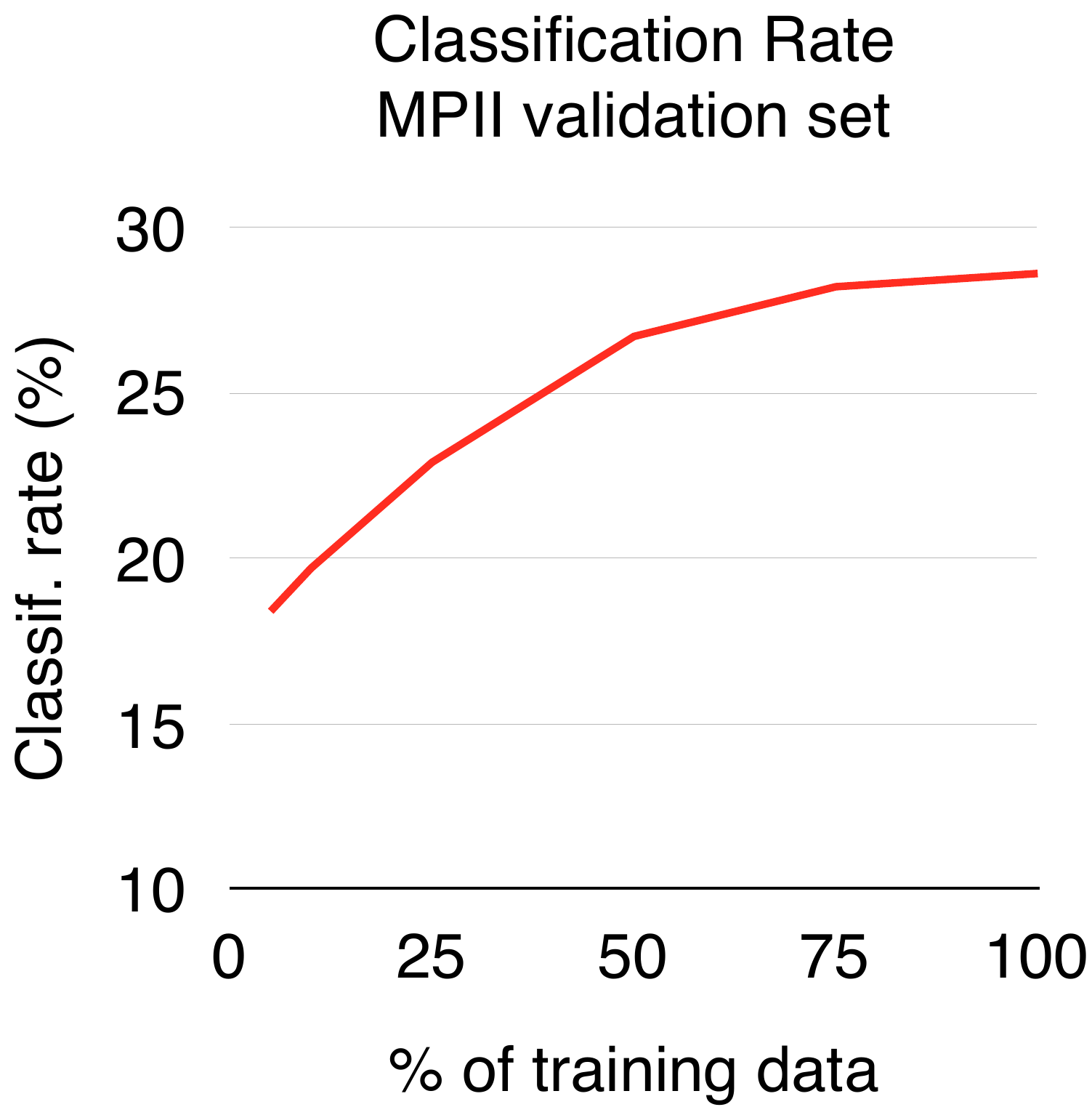}\\
  (a) & (b)
  \end{tabular}
\caption{(a) PCKh@0.5 vs the number K of
  anchor-poses \new{and (b) classification rate for K=200 when varying the percentage of training data randomly picked in the full set 200k images}. Results are reported  
on  MPII validation set   \new{for a model with a VGG backbone  after PPI with $T_{3D}=0.2$ m and $IoU=0.2$}.}
     \label{fig:PCKhvsK} 
 \end{figure}

 \noindent {\bf Dealing with truncation.}  
To deal with truncations by the image boundary, we double the number of clusters by considering also upper-body region proposals.
More precisely, for the $K$ anchor-poses, we adjust the full-body anchor-pose such that only the upper-body covers the candidate box but we still regress the full-body pose. This process allows us to ``hallucinate'' valid full-body poses even when only the upper-body is visible.
At training, we define an upper-body ground-truth box for each annotated pose plus a fully-body ground-truth box when at least one joint from the lower limbs is visible. By this process, we obtain 476k upper-body and 415k full-body poses in our training set. 

 \noindent {\bf Pseudo ground-truth 3D pose and anchor-poses.} 
LCR-Net requires 3D 
ground-truth poses associated with each training image. For MPII, LSPE and MS Coco images, we infer them using the proposed nearest neighbor (NN) search on the annotated joints, see Section~\ref{sub:pseudogt}.   
We consider the CMU MoCap dataset as 3D pose source, as in \cite{RogezS16,IqbalGG16}.
However, both MPII and LSPE datasets present rare poses (\eg, gymnastic) that are absent from this  dataset. 
To cover a wider set of poses, we merged several MoCap datasets available on the internet, such as Pose Prior~\cite{AkhterB15} and HDM05~\cite{HDM05},  and observed a $13\%$ reduction in the matching error, \ie, distance between the query 2D pose and the best match, when using this augmented dataset.
The set of anchor-poses is obtained by running $K$-means on the 3D poses of the extended MoCap dataset.
In Figure~\ref{fig:PCKhvsK}a, we show PCKh when varying the number $K$ of anchor-poses.  
Compared to Human3.6M, the diversity in pose is significantly higher and we found that an optimum number is reached for $K$=200.  We keep $K$=200 anchor-poses for the remaining experiments in the following.

\begin{figure}[htb] 
   \centering 
   \begin{tabular}{cc} 
  \hspace{-0.2cm} \includegraphics[trim={8mm 0mm 18mm 2mm},clip,height=0.21\textwidth]{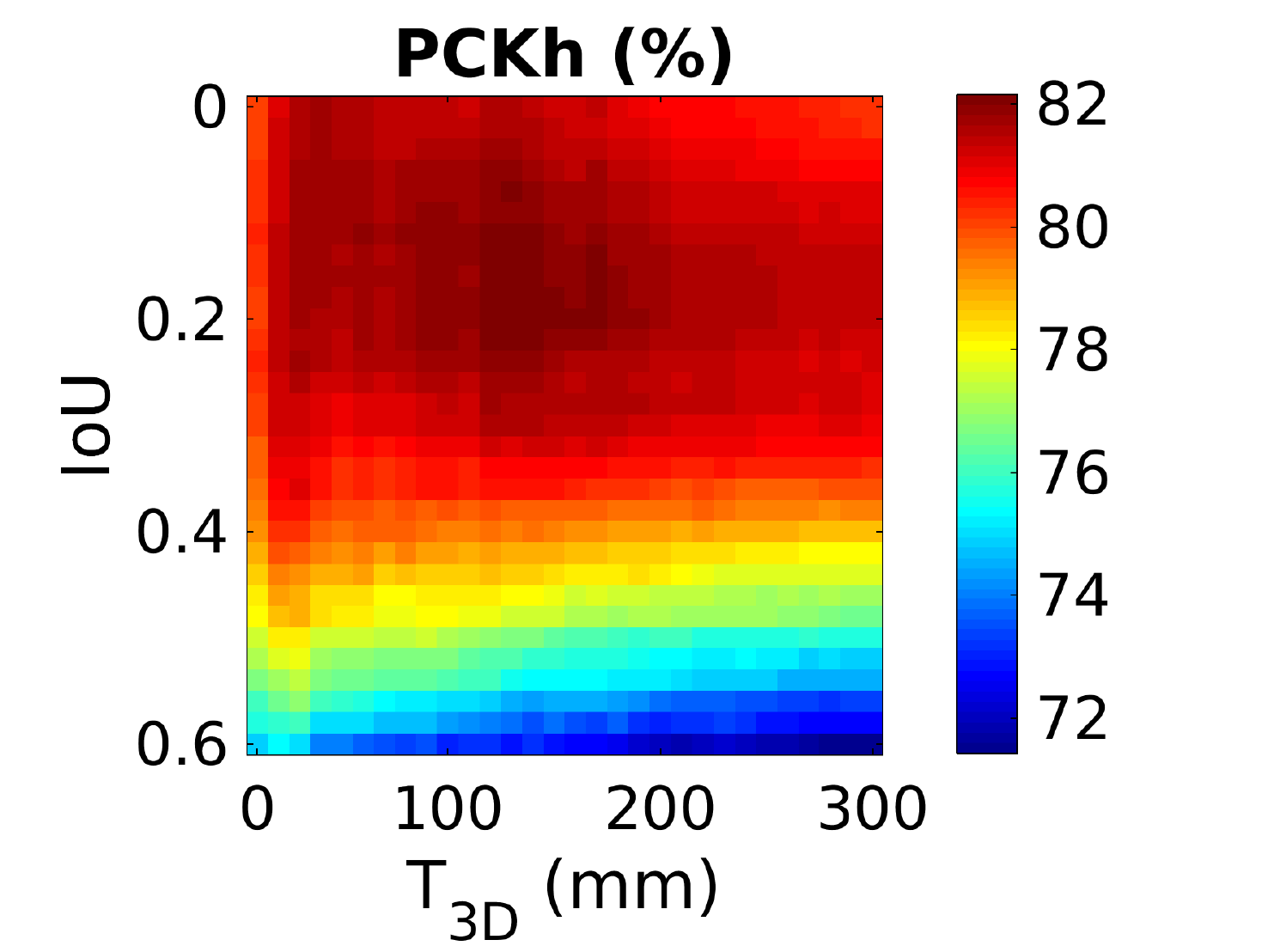}&
  \hspace{-0.2cm} \includegraphics[trim={8mm 0mm 18mm 2mm},clip,height=0.21\textwidth]{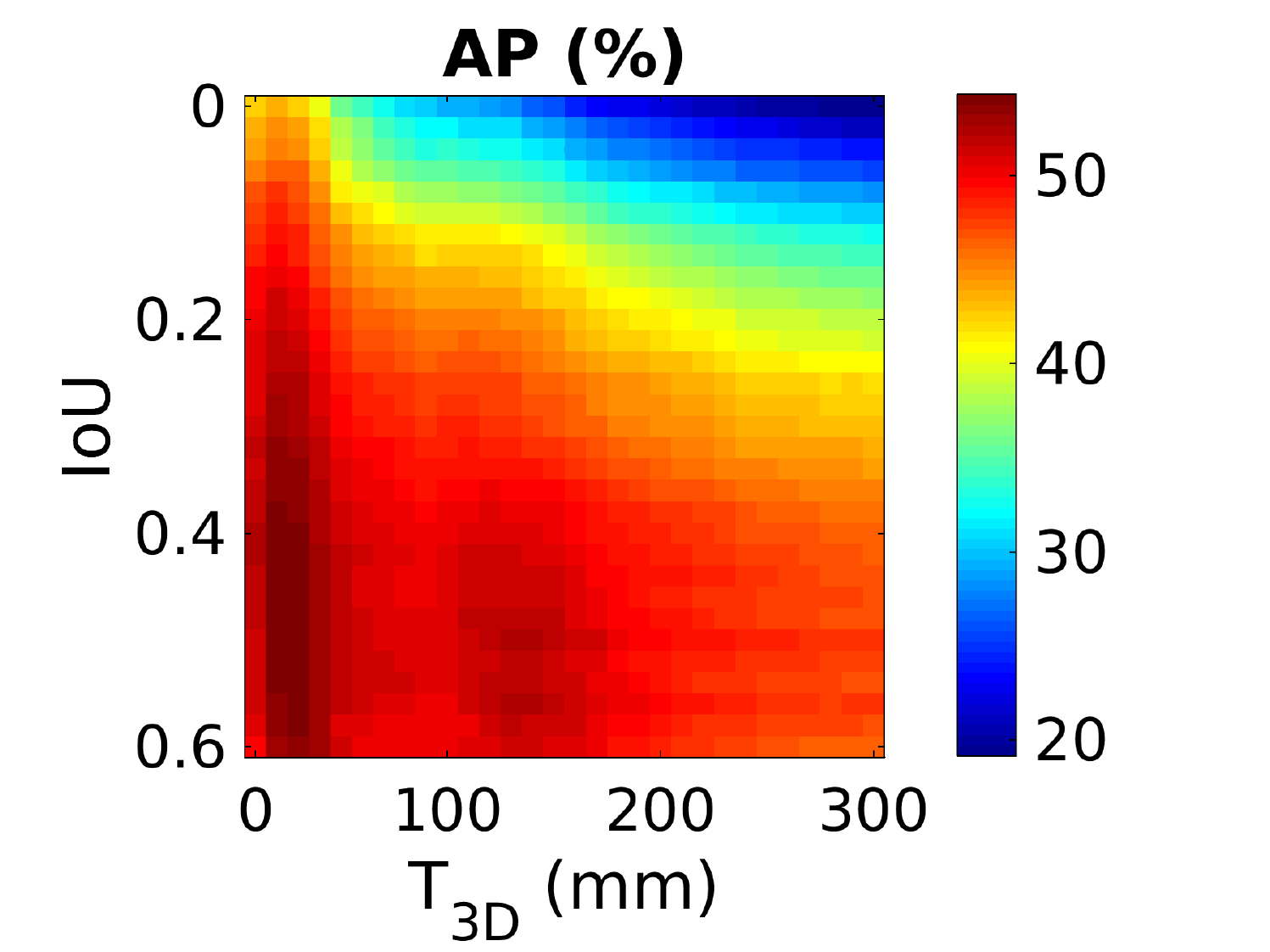}
  \end{tabular}
\caption{Single-person PCKh@0.5 (left) and multi-person AP (right) on MPII validation set when varying IoU and $T_{3D}$ \new{for a model with a VGG backbone}.
}
     \label{fig:plotsPPI} 
 \end{figure}

\noindent {\bf Impact of PPI.}  
We  experimentally set $T_{3D}$ to $130$ mm and  IoU to $0.12$ to evaluate PCKh \new{when using a VGG backbone}, see Figure~\ref{fig:plotsPPI}. As
expected, the IoU threshold has greater impact on multi-person AP than
on single person PCKh: with a small IoU, pose proposals corresponding
to different persons with a high spatial overlap in 2D can be
accidentally merged if they correspond to similar 3D poses. A group of people moving together (\eg dancers) is a typical failure
case. With $T_{3D}$=130 mm and IoU=0.12, we obtain PCKh@0.5=82.16\% \new{when using a VGG backbone. The best multi-person performance with this architecture (AP=54.31\%) is obtained with $T_{3D}$=30 mm and IoU=0.54. With a ResNet50 backbone, we reached PCKh@0.5=87\% and AP=61.7\% with lower values $T_{3D}$=40 mm and IoU=0.02, indicating that the estimated poses are more accurate.}
\begin{table}[htb]
\centering
\resizebox{\linewidth}{!}{
\begin{tabular}{lcc}
\toprule
    &   NMS   &  PPI  \\
\midrule
LCR-Net~\cite{RogezWS17}    &69.87& 75.21 \\ 
 \hspace{0.5cm} + MS-Coco training set   &  74.84&  79.95  \\ 
 \hspace{0.75cm}  + Synthetic data    &76.30&  80.79     \\
\hspace{1cm}  + RoI align    &  78.36 &81.32\\
\hspace{1.25cm}  + iterative estimation     & 78.76& 81.78 \\
\hspace{1.5cm}  + rescoring  \new{(\textbf{LCR-Net+})  }  & 80.30& 82.16  \\
\hspace{1.25cm} \new{+ ResNet50 backbone }   &  \new{85.31 }&   \new{86.87} \\
\hspace{1.5cm}  \new{+ rescoring  (\textbf{LCR-Net++})  }& \new{\bf{85.73}}& \new{\bf{87.00}  } \\
\bottomrule
\end{tabular}
}
\vspace{1mm}
\caption{Ablative analysis on MPII validation set.  We evaluate the PCKh@0.5 (\%) of our architecture when adding the different modifications introduced in this work compared to the  version of LCR-Net published in~\cite{RogezWS17} with a RoI pooling layer and trained on MPII+LSPE+Human3.6M images. 
For each tested model, the PCKh@0.5 (\%) is reported for NMS and after PPI \new{(with $T_{3D}$=130 mm \& IoU=0.12 for VGG and $T_{3D}$=40 mm \& IoU=0.02 for ResNet50).}}
\label{tab:MPII_ablative}
\end{table}

  \begin{figure*}[htb]    
       \centering   
 \hspace{-5mm}\includegraphics[trim={0mm 0mm 0mm 75mm},clip,width=\textwidth]{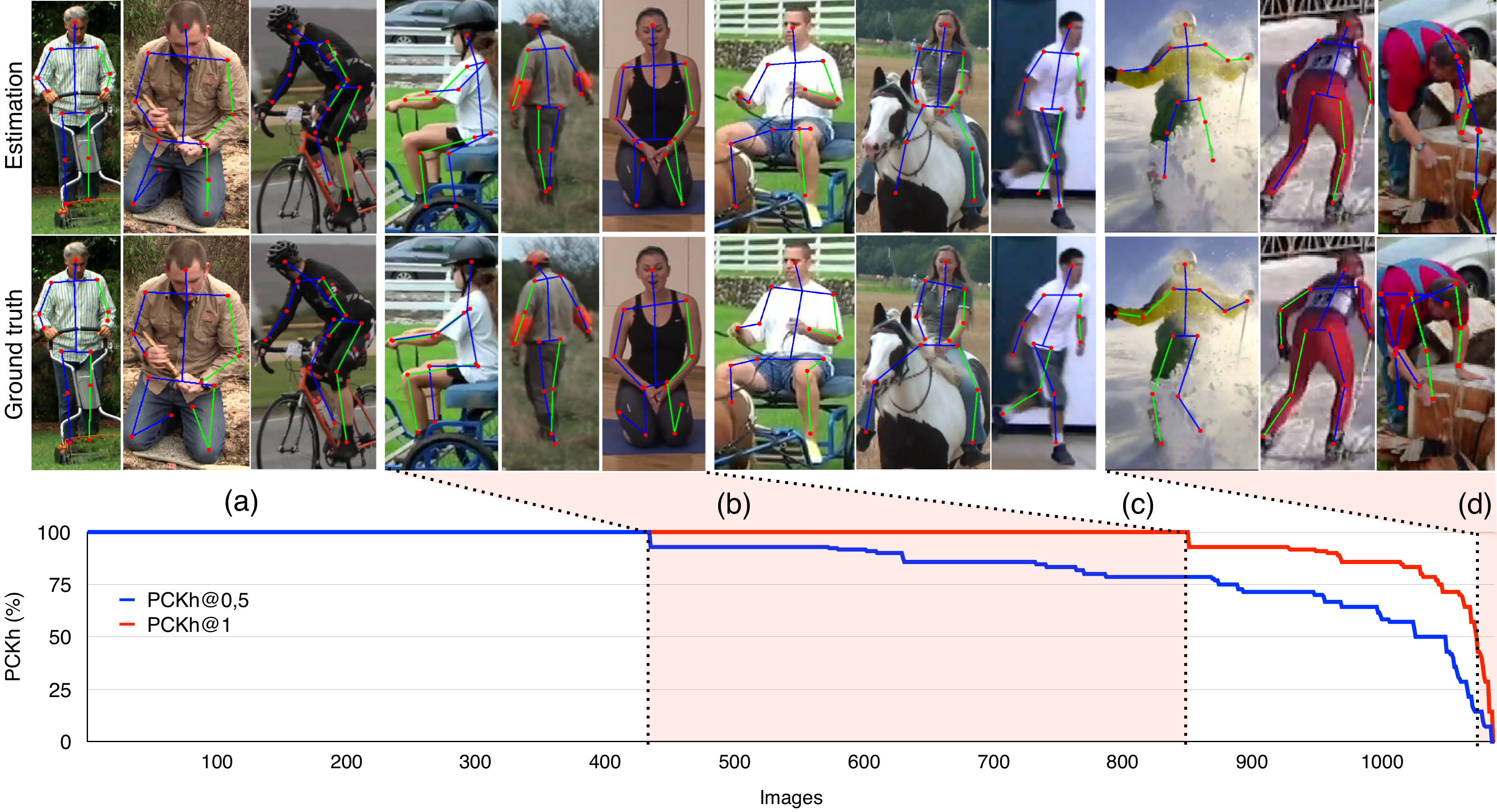}  
\caption{Qualitative analysis on MPII validation set. The average ``per-pose'' PCKh@0.5  and PCKh@1 is represented (bottom) when ordering the poses with respect to PCKh score. \new{This  visualizes (from left to right) the poses that are (a) perfectly recognized (40\% of the poses),  (b) correct but imprecise (38.2\%), (c) partly incorrect (20.5\%) and (d) miss-detected (1.3\%). For each category, we show 3 examples of  estimated poses (top) and the corresponding ground truth annotations (middle).  Note that on average, PCKh@0.5=87.0\% and PCKh@1=96.2\% (with a ResNet50 backbone)}.
}
     \label{fig:plotsPCKh} 
 \end{figure*}  
 \noindent {\bf Ablative analysis.} 
We provide an ablative analysis for single person pose estimation on the MPII validation set  in Table~\ref{tab:MPII_ablative}. 
On our validation set, the initial version of LCR-Net\cite{RogezWS17} (trained on MPII, LSPE and Human3.6M) obtains  75.21\% for a standard PCKh@0.5.  We can see that PPI (with $T_{3D}$=130 mm and IoU=0.12) improves with respect to NMS by 5.34\%.
When adding annotated images from 
MS-Coco~\cite{LinMBHPRDZ14}, \ie, approximately doubling the size of the
training set,  a significant improvement in performance is obtained, PCKh@0.5=79.95\% on the
validation set.  This confirms that LCR-Net requires a large
amount of training data. While using additional synthetic data had a strong impact on the performance in the experiments on the Human3.6M dataset, the improvement is less substantial when evaluating on MPII validation set (+0.84\%). A possible explanation is that generating useful synthetic data is much harder in-the-wild than in the controlled Human3.6M scenario where no occlusions, no object manipulations and no multi-persons scenes are observed. \new{In Figure~\ref{fig:PCKhvsK}b, we show the impact of more ground truth examples on classification performance for K=200 by retraining LCR-Net on increasing random subsets of the 200k training images. Convergence is reached when using the full set and adding more training data would not impact much the performance.
In return, we can see in Table~\ref{tab:MPII_ablative} }that the RoI alignment greatly improves the quality of the region features, since the gap between NMS  and PPI results decreases from 4.49\% to 2.96\%. The iterative refinement improves the performance by another 0.46\%. The rescoring of the pose proposals (Equation~\ref{eqn:regscore} in Section~\ref{sect:PPI}) helps to improve the NMS estimates by 1.54\% but has a marginal influence on PPI results (+0.38\%).  \new{Finally, using a ResNet50 backbone increases the learning capacity of our architecture and considerably boosts the performance, reaching $87\%$ after rescoring. LCR-Net++ produces significantly more accurate pose proposals than its initial version and the PPI post-processing stage still improves over the simpler NMS but to a lesser extent (1.27\%).}

 \noindent {\bf Impact of  regression target.} 
 Since 2D and 3D poses are regressed together, inaccurate 3D annotations could negatively impact 2D pose estimation. 
To evaluate the effect of  the pseudo  ground-truth on 2D performance, we train a version of the architecture  to predict the 2D poses only (see Table~\ref{tab:MPII_analysis}). 
We observe a decrease of the NMS performance obtaining a PCKh of 74.61\% compared to 76.30\% with 2D+3D regression. Adding the 3D pose regression actually helps to improve the performance in 2D. \new{Regressing also the 3D poses allows to learn better features for the task of 2D pose estimation. In a similar spirit, many works in the literature have shown that multi-task learning is beneficial for each single task.}
Finally, we evaluate PCKh when only considering full-body classes and observed a lower performance after NMS and PPI, validating that adding upper-body classes to the full-body classes improves performance on the MPII validation set.

  \begin{table}[tb]
\centering
\resizebox{\linewidth}{!}{
\begin{tabular}{lcc}
\toprule
    &   NMS   &  PPI  \\
\midrule
Baseline (\cite{RogezWS17} + MS-Coco train + Synth)\qquad\qquad\qquad\qquad &  {\bf 76.30}& {\bf 80.79} \\  
\midrule
Regressing 2D pose only     & 74.61  &  - \\
Using full-body classes only  & 74.42 & 78.40  \\
\bottomrule
\end{tabular}
}
\vspace{1mm}
\caption{Additional analysis on MPII validation set.  We evaluate the performance  of  LCR-Net  (a) when predicting only the 2D poses and (b) when using only full-body classes, \ie, no upper-body classes. 
For each tested model, the PCKh@0.5 (\%) is reported for NMS and after PPI with $T_{3D}$=130 mm and IoU=0.12. }
\label{tab:MPII_analysis}
\end{table}

 \begin{table}
\centering
\resizebox{\linewidth}{!}{
\begin{tabular}{lcccccccc }
\toprule
 &Head & Shoulder & Elbow & Wrist & Hip & Knee  & Ankle & PCKh \\
 \midrule
Yang~\etal~\cite{YangLOLW17}& 98.5  & 96.7  & 92.5  & 88.7  & 91.1  & 88.6 & 86.0 & 92.0  \\
Chu~\etal~\cite{ChuYOMYW17}& 98.5  & 96.3  & 91.9  & 88.1  & 90.6  & 88.0 & 85.0 & 91.5 \\
Newell~\etal~\cite{NewellYD16}& 98.2  & 96.3  & 91.2  & 87.1  & 90.1  & 87.4 & 83.6 & 90.9 \\
Bulat~\etal~\cite{BulatT16}& 97.9  & 95.1  & 89.9  & 85.3  & 89.4  & 85.7 & 81.7 & 89.7 \\
Wei~\etal~\cite{WeiRKS16}& 97.8  & 95.0  & 88.7  & 84.0  & 88.4  & 82.8 & 79.4 & 88.5  \\
\new{\bf{LCR-Net++}}& 93.5  & 94.6  & 88.4  & 80.9  & 88.2  & 80.8 & 71.4 & 86.1\\
Gkioxary~\etal~\cite{GkioxariTJ16}& 96.2  & 93.1  & 86.7  & 82.1  & 85.2  & 81.4 & 74.1 & 86.1 \\
Pishchulin~\etal~\cite{PishchulinITAAG15}& 94.1  & 90.2  & 83.4  & 77.3  & 82.6  & 75.7 & 68.6 & 82.4  \\
Hu\&Ramanan~\cite{HuR16}& 95.0  & 91.6  & 83.0  & 76.6  & 81.9  & 74.5 & 69.5 & 82.4  \\
Carreira~\etal~\cite{CarreiraAFM16}& 95.7  & 91.7  & 81.7  & 72.4  & 82.8  & 73.2 & 66.4 & 81.3 \\
Tompson~\etal~\cite{TompsonJLB14}& 95.8  & 90.3  & 80.5  & 74.3  & 77.6  & 69.7 & 62.8 & 79.6  \\
\bottomrule
\end{tabular}
}
\vspace{2mm}
\caption{\new{2D pose estimation results on single-person MPII test set compared to state-of-the-art 2D methods. }
}
\label{tab:pose2Dsingle} 
\end{table}

\begin{figure*}[htb]    
   \begin{tabular}{ccc}
 \hspace{-3mm}\includegraphics[height=0.4\textwidth]{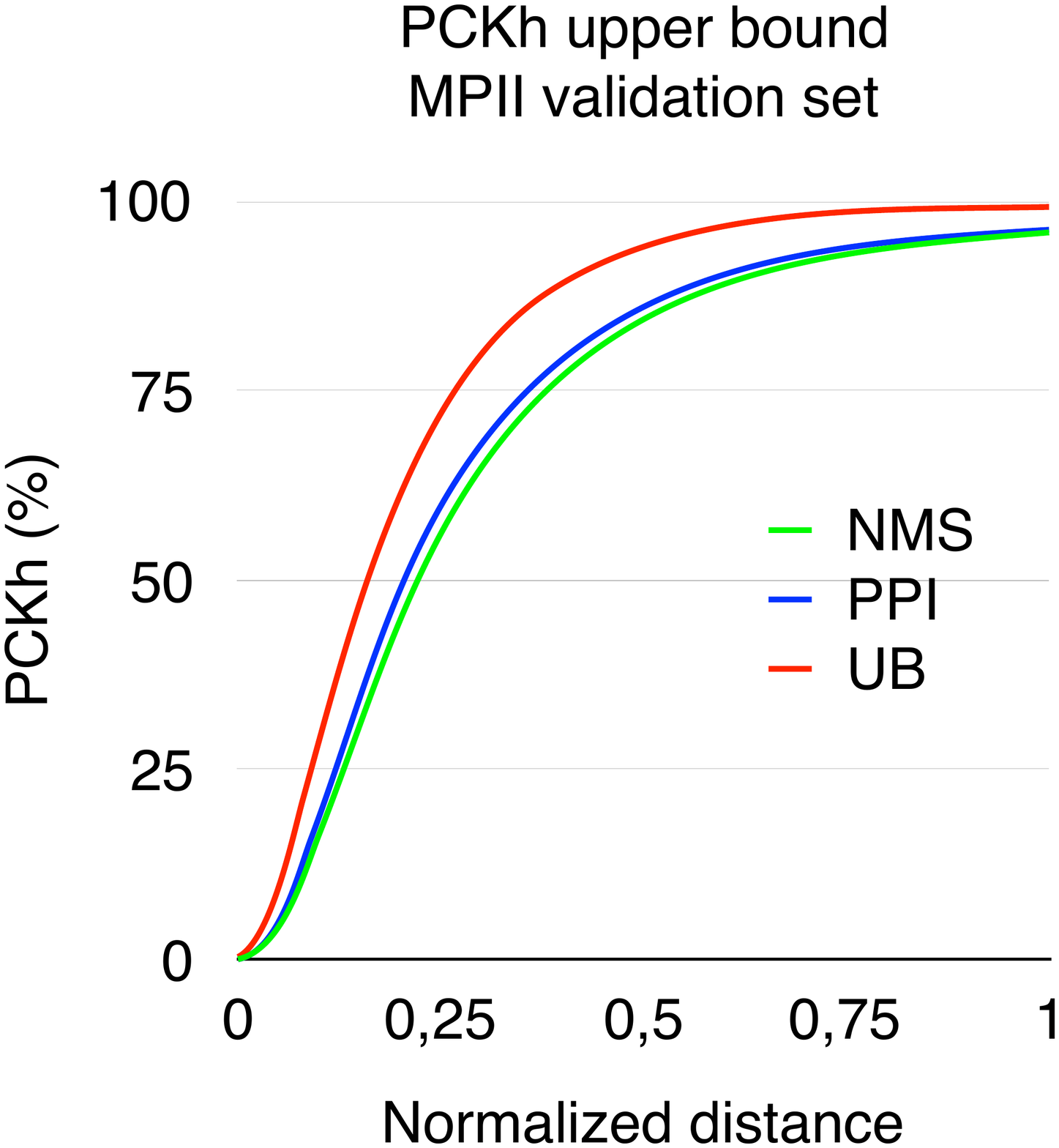} &
 \includegraphics[height=0.4\textwidth]{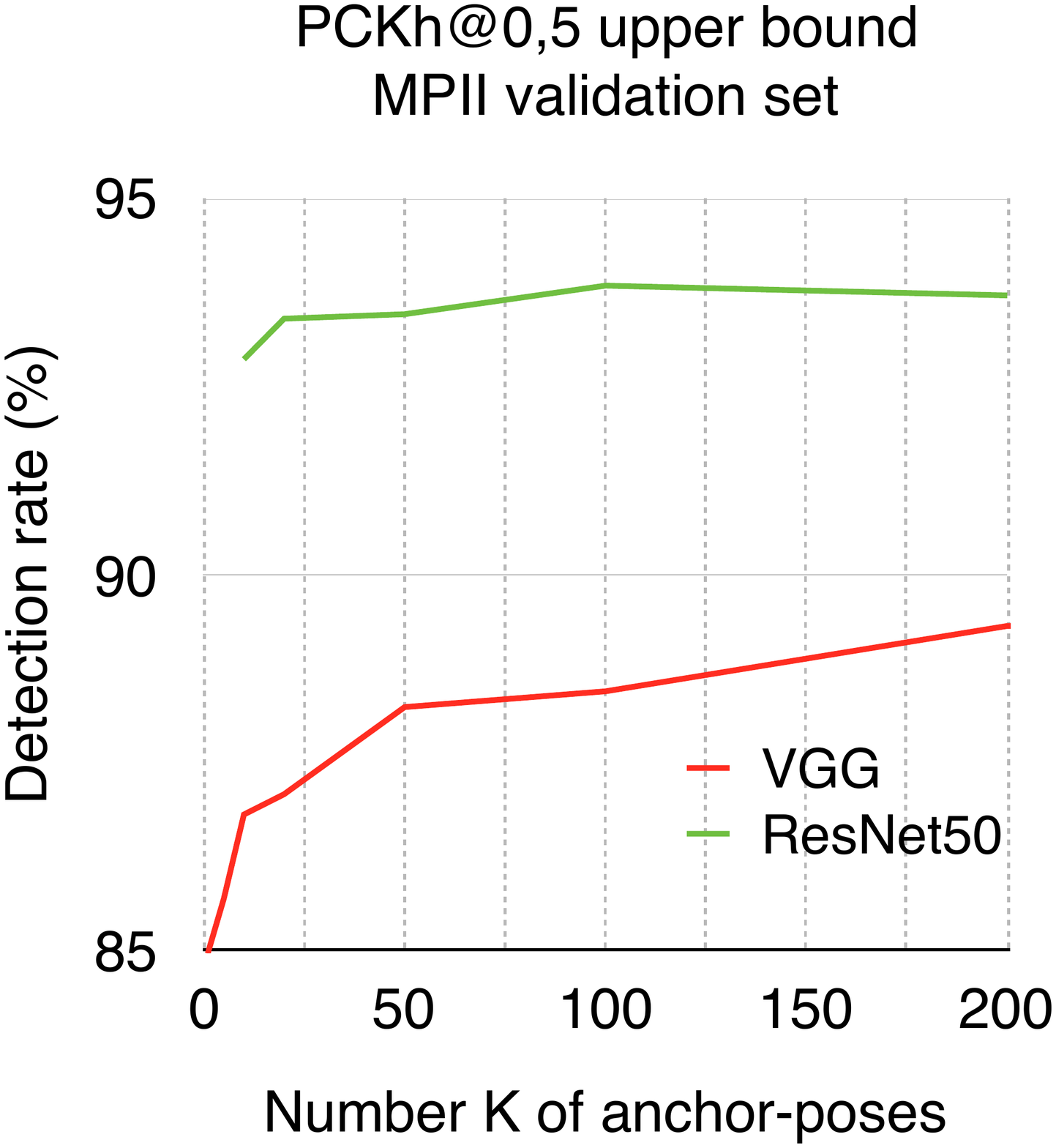} &
  \includegraphics[height=0.4\textwidth]{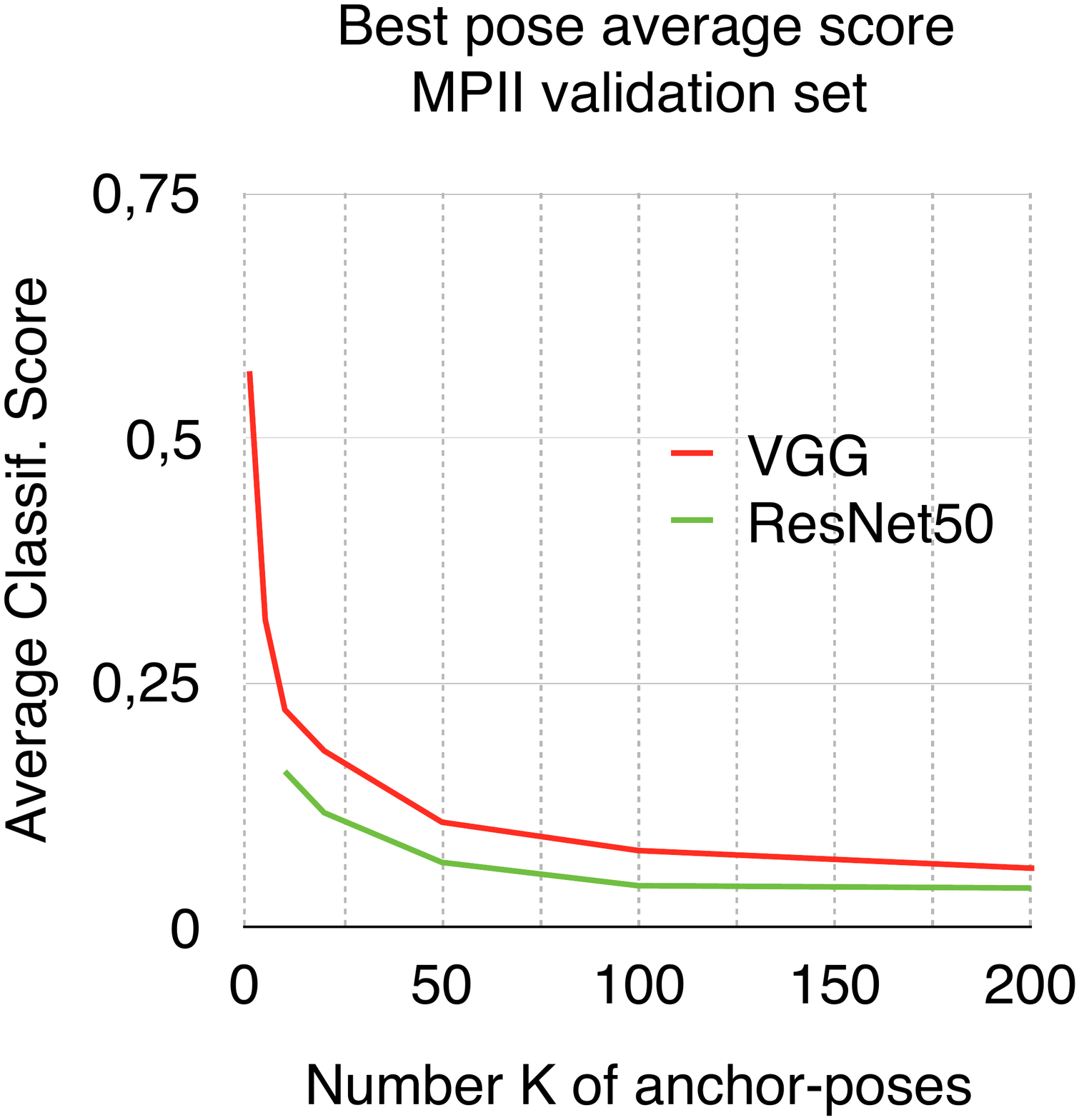}   \vspace{-7mm}\\
    (a) & (b) & (c) 
 \end{tabular} \vspace{-3mm}
\caption{Upper bound on MPII validation set. (a)~Detection rate with respect to the normalized distance in PCKh computation for PPI, NMS and the Upper bound (UB), \ie, taking the pose proposal closest to ground-truth pose. Results are reported for K=200 anchor-poses. (b)~Upper bound  of the PCKh@0.5 when varying K, the number of anchor-poses. (c)~Average score of the pose proposals used to compute the upper bound. \new{In (b) and (c), we report the results obtained when considering a VGG16 or a ResNet50 backbone in LCR-Net.}
}
     \label{fig:plotsUB} 
 \end{figure*} 
   \begin{figure*}[htb]
\centering
\includegraphics[width=\linewidth]{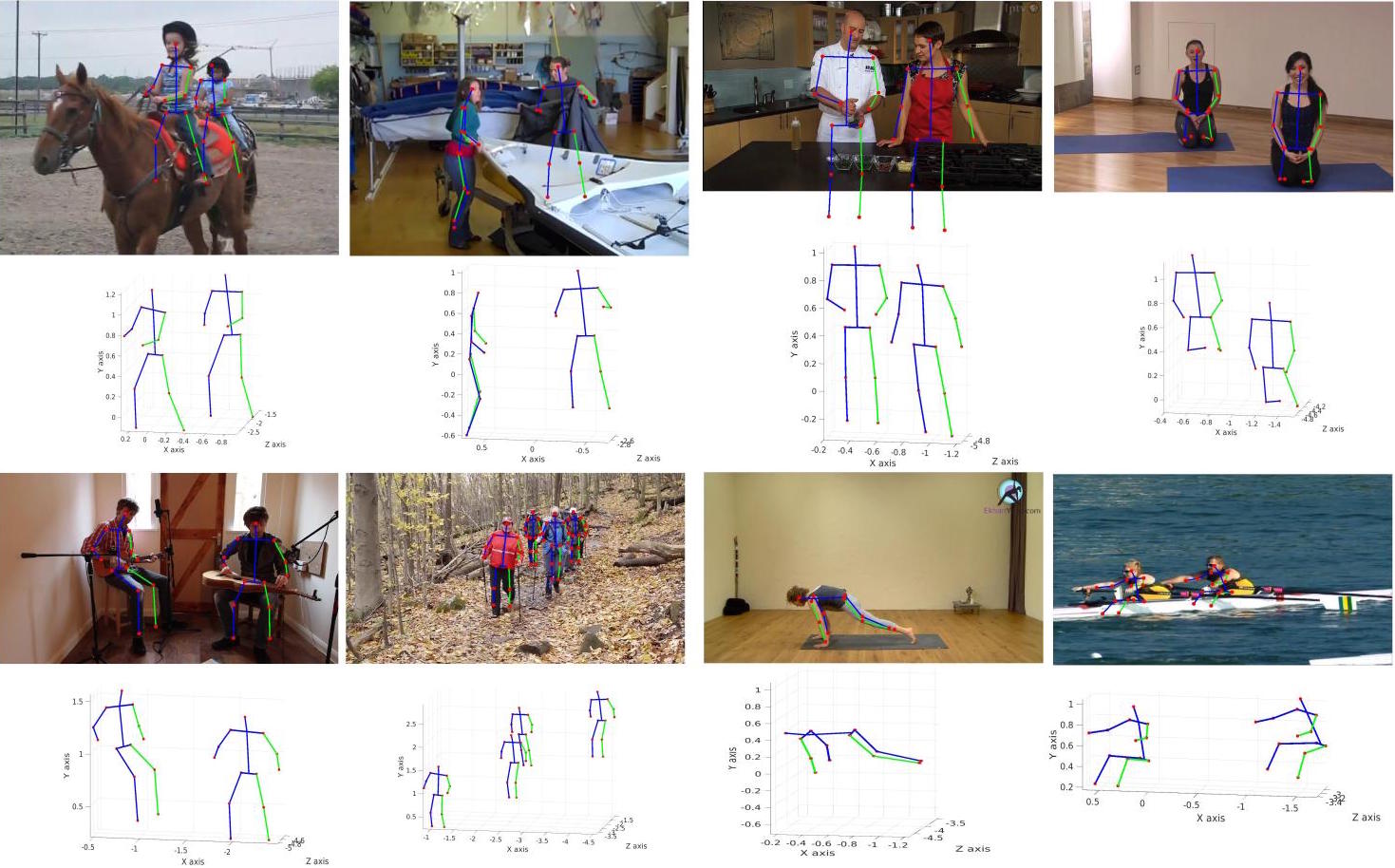}
\caption{Qualitative examples. LCR-Net outputs multiple 2D and 3D poses, the 3D poses being expressed in a camera reference system
  centered on the torso.
  To represent the 3D poses in a common coordinate system,
  we find for each of them the appropriate 3D displacements in front of the camera.
This is obtained using a least square minimization of the reprojection error, \ie, the distance between 2D pose and reprojected 3D pose. When the camera is unknown, hypothesizing an orthographic camera
leads to acceptable qualitative results as shown in these examples.  
}
\label{fig:mpii_res}
\end{figure*}

 \noindent {\bf Detailed analysis.} 
 \new{While we outperform the state of the art in 3D human pose estimation in a controlled environment,
our 2D performance on real images is comparable to other recent competing methods but below the state of the art on the MPII test set, as reported in Table~\ref{tab:pose2Dsingle}}.  Note that in
contrast to most other approaches, our holistic method also gives an
estimation of the occluded joints that is not evaluated. 
Figure~\ref{fig:plotsPCKh} shows the  ``per pose'' PCKh on the validation set for PCKh@0.5  and PCKh@1. The poses are ordered with respect to PCKh score. 
 \new{ We can see (from left to right) that $78.2\%$ of the poses are globally correct ($40\%$ of the poses are perfectly recognized and $38.2\%$ are simply imprecise) while $20.5\%$ are partly incorrect, e.g. a limb is poorly estimated, and $1.3\%$ of the poses are miss-detected, \ie, PCKh@1$\leq$50\%.} These misdetections are often due to a right-left inversion in the estimation (or in the ground-truth annotations) leading to very poor PCKh scores as visualized in the examples. We can see on Figure~\ref{fig:plotsUB}a  that PCKh@1 approaches $95\%$. Although globally correct, our pose estimations can lack precision on the limb extremities resulting in lower PCKh score in 2D.
 One explanation is that we use a fully-connected layer for the regression. This could be improved by using fully 
convolutional architecture with deconvolution or upsampling~\cite{NewellYD16}.  
Another possible explanation is that the pose proposals are not correctly scored.
On Figure~\ref{fig:plotsUB}a, we can see that if we compute the upper bound on the PCKh, \ie, computed with the closest pose proposals from ground-truth annotations, we can obtain greater performances: PCKh@0.5=94  and PCKh@1=99.2. This indicates that  the classification score is not always representative of the quality of the regressed 2D and 3D poses. 
Some high scoring poses can in fact be imprecise while others with lower scores are more accurate. In Figure~\ref{fig:plotsUB}b, we show this upper bound of the PCKh@0.5 when varying the number $K$ of anchor-poses. Adding more anchor-poses helps generate better pose proposals but their score decreases when augmenting $K$ as shown in Figure~\ref{fig:plotsUB}c.  The
anchor-poses become probably too similar and harder to distinguish, resulting in ambiguities in the classification. Another reasons for this observation could be the amount of training data available for each class that also decreases when increasing $K$. We proposed a rescoring function that helps to improve both NMS and PPI performances but a better scoring function should be investigated in future work.

 \noindent {\bf Multi-person pose detection.} 
For multi-person evaluation, our validation set contains 209 groups of multiple people in 187 images.
We follow the standard protocol and evaluate AP averaged over joints. 
 \new{We obtain $54.3\%$ for a standard mAP@0.5 with LCR-Net+ and $61.7 \%$ with LCR-Net++. We found that considering only head and torso keypoints (hips and shoulders) to define the 2D bounding box employed in the IoU computation of PPI helps avoiding unwanted merges in case of people very close to each other, reaching mAP@0.5=$64.9\%$.
Our performance on the test set is only $47.9\%$ (see Table~\ref{tab:pose2Dmulti})}. \new{This is below state-of-the-art, e.g., \cite{NewellHD17} reports mAP@0.5=77.5\%. Our approach fails in very crowded scenes and when the overlap between people is too important, as it is often the case in the test set. Note that we also estimate the 3D poses unlike all reported approaches who only focus on 2D pose estimation. Examples of pose detections are shown in Figure~\ref{fig:mpii_res}.
Our method can detect multiple people even if they overlap ($2^{nd}$ row, $2^{nd}$ column).
It can also tackles unusual poses (top right) and truncations (top row, $3^{rd}$ column). }
 \begin{table}
\centering
\resizebox{\linewidth}{!}{
\begin{tabular}{lcccccccc }
\toprule
 &Head & Shoulder & Elbow & Wrist & Hip & Knee  & Ankle & Total\\
  \midrule
Newell~\etal~\cite{NewellHD17}& 92.1  & 89.3  & 78.9  & 69.8  & 76.2  & 71.6 & 64.7 & 77.5 \\
Cao~\etal~\cite{CaoSWS17}& 91.2  & 87.6  & 77.7  & 66.8  & 75.4  & 68.9 & 61.7 & 75.6 \\
Insafutdinov~\etal~\cite{InsafutdinovAPT17}& 88.8  & 87.0  & 75.9  & 64.9  & 74.2  & 68.8 & 60.5 & 74.3 \\
Insafutdinov~\etal~\cite{InsafutdinovPAAS16}& 78.4  & 72.5  & 60.2  & 51.0  & 57.2  & 52.0 & 45.4 & 59.5 \\
\new{\bf{LCR-Net++}}& 59.0  & 60.5  & 50.8  & 39.5  & 51.2  & 42.8 & 31.7 & 47.9 \\
Iqbal\&Gall~\cite{IqbalG16}& 58.4  & 53.9  & 44.5  & 35.0  & 42.2  & 36.7 & 31.1 & 43.1 \\
\bottomrule
\end{tabular}
}
\vspace{2mm}
\caption{\new{2D pose estimation results on multi-person MPII test set compared to state-of-the-art 2D methods. }
}
\label{tab:pose2Dmulti} 
\end{table}

\subsection{\new{Multi-person 3D pose detection on MuPoTS-3D}}
\label{sub:mupots}

\new{In this section, we evaluate LCR-Net++ on the  Multi-person Pose estimation Test Set in 3D (MuPoTS-3D)~\cite{MehtaSMXSPT18}. This dataset of around 8k frames comprises 20 real-world scenes with ground-truth 3D pose for up to three subjects obtained with a multi-view marker-less MoCap system. 
Mehta~\etal~\cite{MehtaSMXSPT18} introduce occlusion-robust pose-maps which enable full body pose inference for an arbitrary number of people even under partial occlusions. To train their approach, they employ a large scale training data set of real and composited images with ground truth 3D poses also obtained with a multi-view MoCap system. To better fit the 14-joint skeleton model measured by their MoCap system, we finetuned our LCR-Net++ architecture on the same training data. As in \cite{MehtaSMXSPT18}, we report the 3DPCK (percentage of joint prediction within a 15cm ball centred on ground-truth) per sequence, averaged over the subjects for which ground truth is available. Results are reported in Table~\ref{tab:MuPoTS}. We establish a new state-of-the-art performance of $70.6\%$ ($74\%$ when evaluating only on well-detected persons) compared to $65.0\%$ for Mehta \etal~\cite{MehtaSMXSPT18} ($69.8\%$ ignoring mis-detections).}

\begin{table*}[htb]
\resizebox{\linewidth}{!}{
\begin{tabular}{llccccccccccccccccccccc}
\toprule
& Method   & TS1& TS2& TS3& TS4& TS5& TS6& TS7& TS8& TS9& TS10& TS11& TS12& TS13& TS14& TS15& TS16& TS17& TS18& TS19& TS20&  Avg. \\
\midrule
\midrule
&LCR-Net~\cite{RogezWS17} & 67.7 & 49.8 & 53.4 & 59.1 & 67.5 & 22.8  &43.7 & 49.9 & 31.1  &78.1   &50.2  & 51.0  & 51.6 & 49.3  & 56.2&   66.5  & 65.2 &  62.9 &  66.1 &  59.1&
53.8\\ 
(a)&Mehta~\etal~\cite{MehtaSMXSPT18} & 81.0&  59.9&  64.4&  62.8 & 68.0 & 30.3& {\bf65.0} & 59.2&  64.1 & 83.9&  67.2&  68.3&  60.6&  {\bf56.5}&  69.9 & 79.4&  {\bf79.6}&  66.1&  66.3 & {\bf63.5}&  65.0\\
&\bf{LCR-Net++} & {\bf87.3}& {\bf61.9}& {\bf67.9}& {\bf74.6}& {\bf78.8}& {\bf48.9}&  58.3 & {\bf59.7}& {\bf78.1}& {\bf89.5}& {\bf69.2}&{\bf73.8}& {\bf66.2}& 56.0& {\bf74.1}& {\bf82.1}& 78.1& {\bf72.6}& {\bf73.1}& 61.0 &  {\bf70.6} \\

\midrule
&LCR-Net~\cite{RogezWS17} & 69.1&67.3&54.6 & 61.7&74.5&25.2&  48.4&63.3& 69.0&78.1&   53.8  & 52.2 &  60.5 &  60.9 &  59.1  & 70.5  & 76.0  & 70.0&77.1&81.4&62.4  \\
(b)&Mehta~\etal~\cite{MehtaSMXSPT18} & 81.0&64.3&64.6 &63.7&73.8&30.3 &{\bf65.1}&60.7 & 64.1&83.9  &{\bf71.5}  &69.6 & 69.0 & {\bf69.6}  &71.1 & 82.9 & 79.6 & 72.2&{\bf76.2}&{\bf85.9}  & 69.8\\
 
&\bf{LCR-Net++} & {\bf88.0}& {\bf73.3}& {\bf67.9}& {\bf74.6}& {\bf81.8}& {\bf50.1}& 60.6& {\bf60.8}& {\bf78.2}& {\bf89.5}& 70.8& {\bf74.4}& {\bf72.8}& 64.5& {\bf74.2}& {\bf84.9}& {\bf85.2}& {\bf78.4}& 75.8& 74.4 &  {\bf74.0} \\
\bottomrule
\end{tabular}
}
\vspace{1mm}
\caption{\new{Sequence-wise  evaluation  of  our  method and \cite{MehtaSMXSPT18} on their multi-person  3D  pose  test set MuPoTS-3D. As in~\cite{MehtaSMXSPT18}, we report both (a) the overall detection accuracy in $\%$ (\ie, 3DPCK within a 15 cm ball ), and (b) the accuracy only for person annotations matched to a prediction.}}
\label{tab:MuPoTS}
\end{table*}

\section{Conclusion}

This paper introduces a Localization-Classification-Regression network
(LCR-Net) for joint 2D and 3D human pose detection in natural
images. We demonstrate the benefit of an end-to-end architecture
which relies on pose proposals that are hypothesized at different
locations in the image, scored by classification and refined  by regression. The
final pose estimation is obtained by integrating over neighboring pose
hypotheses. We outperform the state of the art in 3D pose estimation
in controlled environments and show promising results on real
images. 

The upper bound performance shows that there is room for improvement and that a considerable boost could be obtained by  adequately scoring the pose proposals. Our first attempt at rescoring them has shown encouraging results in that direction.
 Another line of improvement concerns the training data. In this work, we proposed a solution to automatically annotate 2D images with ``pseudo'' ground-truth 3D poses. Our ongoing research indicates that  better 3D and 2D performances could be obtained with LCR-Net if more accurate real-world training data was available, e.g. through manual curation.

\vspace{2mm}
 \noindent {\bf Acknowledgements.} 
This work was supported by ERC advanced grant Allegro and an Amazon Academic Research Award.
We  thank NVIDIA  for donating the GPUs
used for this research.

\bibliographystyle{IEEEtran}
\bibliography{biblio}

\begin{thebibliography}{10}
\providecommand{\url}[1]{#1}
\csname url@samestyle\endcsname
\providecommand{\newblock}{\relax}
\providecommand{\bibinfo}[2]{#2}
\providecommand{\BIBentrySTDinterwordspacing}{\spaceskip=0pt\relax}
\providecommand{\BIBentryALTinterwordstretchfactor}{4}
\providecommand{\BIBentryALTinterwordspacing}{\spaceskip=\fontdimen2\font plus
\BIBentryALTinterwordstretchfactor\fontdimen3\font minus
  \fontdimen4\font\relax}
\providecommand{\BIBforeignlanguage}[2]{{%
\expandafter\ifx\csname l@#1\endcsname\relax
\typeout{** WARNING: IEEEtran.bst: No hyphenation pattern has been}%
\typeout{** loaded for the language `#1'. Using the pattern for}%
\typeout{** the default language instead.}%
\else
\language=\csname l@#1\endcsname
\fi
#2}}
\providecommand{\BIBdecl}{\relax}
\BIBdecl

\bibitem{CaoSWS17}
Z.~Cao, T.~Simon, S.~Wei, and Y.~Sheikh, ``Realtime multi-person {2D} pose
  estimation using part affinity fields,'' in \emph{{CVPR}}, 2017.

\bibitem{BulatT16}
A.~Bulat and G.~Tzimiropoulos, ``Human pose estimation via convolutional part
  heatmap regression,'' in \emph{ECCV}, 2016.

\bibitem{NewellYD16}
A.~Newell, K.~Yang, and J.~Deng, ``Stacked hourglass networks for human pose
  estimation,'' in \emph{ECCV}, 2016.

\bibitem{SigalBB10}
L.~Sigal, A.~O. Balan, and M.~J. Black, ``{HumanEva}: Synchronized video and
  motion capture dataset and baseline algorithm for evaluation of articulated
  human motion,'' \emph{IJCV}, 2010.

\bibitem{IonescuPOS14}
C.~Ionescu, D.~Papava, V.~Olaru, and C.~Sminchisescu, ``Human3.6{M}: Large
  scale datasets and predictive methods for {3D} human sensing in natural
  environments,'' \emph{IEEE Trans. PAMI}, 2014.

\bibitem{CMUposedataset}
``{CMU} motion capture dataset. http://mocap.cs.cmu.edu. the database was
  created with funding from {NSF} eia-0196217.''

\bibitem{ChenWLSW16}
W.~Chen, H.~Wang, Y.~Li, H.~Su, Z.~Wang, C.~Tu, D.~Lischinski, D.~Cohen-Or, and
  B.~Chen, ``Synthesizing training images for boosting human {3D} pose
  estimation,'' in \emph{3DV}, 2016.

\bibitem{RogezS16}
G.~Rogez and C.~Schmid, ``Mo{C}ap-guided data augmentation for {3D} pose
  estimation in the wild,'' in \emph{NIPS}, 2016.

\bibitem{FasterRCNN}
S.~Ren, K.~He, R.~Girshick, and J.~Sun, ``Faster {R-CNN}: Towards real-time
  object detection with region proposal networks,'' in \emph{NIPS}, 2015.

\bibitem{ParkHK16}
S.~Park, J.~Hwang, and N.~Kwak, ``3{D} human pose estimation using
  convolutional neural networks with 2{D} pose information,'' in \emph{ECCV
  Workshop}, 2016.

\bibitem{TekinRLF16}
B.~Tekin, A.~Rozantsev, V.~Lepetit, and P.~Fua, ``Direct prediction of {3D}
  body poses from motion compensated sequences,'' in \emph{CVPR}, 2016.

\bibitem{LiC14}
S.~Li and A.~B. Chan, ``{3D} human pose estimation from monocular images with
  deep convolutional neural network,'' in \emph{ACCV}, 2014.

\bibitem{ToshevS14_DeepPose}
A.~Toshev and C.~Szegedy, ``Deep{P}ose: Human pose estimation via deep neural
  networks,'' in \emph{CVPR}, 2014.

\bibitem{RogezWS17}
G.~Rogez, P.~Weinzaepfel, and C.~Schmid, ``{LCR-Net:
  Localization-Classification-Regression for human pose},'' in \emph{CVPR},
  2017.

\bibitem{he2017mask}
K.~He, G.~Gkioxari, P.~Doll{\'a}r, and R.~Girshick, ``Mask {R-CNN},'' in
  \emph{ICCV}, 2017.

\bibitem{HeZRS16}
K.~He, X.~Zhang, S.~Ren, and J.~Sun, ``Deep residual learning for image
  recognition,'' in \emph{{CVPR}}, 2016.

\bibitem{ChenY14}
X.~Chen and A.~L. Yuille, ``Articulated pose estimation by a graphical model
  with image dependent pairwise relations,'' in \emph{NIPS}, 2014.

\bibitem{FanZLW15}
X.~Fan, K.~Zheng, Y.~Lin, and S.~Wang, ``Combining local appearance and
  holistic view: Dual-source deep neural networks for human pose estimation,''
  in \emph{CVPR}, 2015.

\bibitem{OuyangCW14}
W.~Ouyang, X.~Chu, and X.~Wang, ``Multi-source deep learning for human pose
  estimation,'' in \emph{CVPR}, 2014.

\bibitem{TompsonJLB14}
J.~J. Tompson, A.~Jain, Y.~LeCun, and C.~Bregler, ``Joint training of a
  convolutional network and a graphical model for human pose estimation,'' in
  \emph{NIPS}, 2014.

\bibitem{WeiRKS16}
S.-E. Wei, V.~Ramakrishna, T.~Kanade, and Y.~Sheikh, ``Convolutional pose
  machines,'' in \emph{CVPR}, 2016.

\bibitem{PapandreouZKTTB17}
G.~Papandreou, T.~Zhu, N.~Kanazawa, A.~Toshev, J.~Tompson, C.~Bregler, and
  K.~Murphy, ``Towards accurate multi-person pose estimation in the wild,'' in
  \emph{{CVPR}}, 2017.

\bibitem{PishchulinITAAG15}
L.~Pishchulin, E.~Insafutdinov, S.~Tang, B.~Andres, M.~Andriluka, P.~V. Gehler,
  and B.~Schiele, ``Deepcut: Joint subset partition and labeling for multi
  person pose estimation,'' \emph{CVPR}, 2016.

\bibitem{AkhterB15}
I.~Akhter and M.~Black, ``Pose-conditioned joint angle limits for {3D} human
  pose reconstruction,'' in \emph{CVPR}, 2015.

\bibitem{FanZZW14}
X.~Fan, K.~Zheng, Y.~Zhou, and S.~Wang, ``Pose locality constrained
  representation for 3{D} human pose reconstruction,'' in \emph{ECCV}, 2014.

\bibitem{BogoKLGRB2016}
F.~Bogo, A.~Kanazawa, C.~Lassner, P.~Gehler, J.~Romero, and M.~J. Black, ``Keep
  it {SMPL}: Automatic estimation of {3D} human pose and shape from a single
  image,'' in \emph{ECCV}, 2016.

\bibitem{Simo-SerraRATM12}
E.~Simo{-}Serra, A.~Ramisa, G.~Aleny{\`{a}}, C.~Torras, and F.~Moreno{-}Noguer,
  ``Single image {3D} human pose estimation from noisy observations,'' in
  \emph{CVPR}, 2012.

\bibitem{WangWLYG14}
C.~Wang, Y.~Wang, Z.~Lin, A.~L. Yuille, and W.~Gao, ``Robust estimation of {3D}
  human poses from a single image,'' in \emph{CVPR}, 2014.

\bibitem{RamakrishnaECCV2012}
V.~Ramakrishna, T.~Kanade, and Y.~A. Sheikh, ``{Reconstructing {3D} human pose
  from {2D} image landmarks},'' in \emph{ECCV}, 2012.

\bibitem{ChenR17}
C.~Chen and D.~Ramanan, ``{3D} human pose estimation = {2D} pose estimation +
  matching,'' in \emph{{CVPR}}, 2017.

\bibitem{Moreno17}
F.~Moreno-Noguer, ``{ {3D} human pose estimation from a single image via
  distance matrix regression},'' in \emph{CVPR}, 2017.

\bibitem{NieWZ17}
B.~Xiaohan~Nie, P.~Wei, and S.-C. Zhu, ``Monocular {3D} human pose estimation
  by predicting depth on joints,'' in \emph{ICCV}, 2017.

\bibitem{MartinezHRL17}
J.~Martinez, R.~Hossain, J.~Romero, and J.~J. Little, ``A simple yet effective
  baseline for {3D} human pose estimation,'' in \emph{ICCV}, 2017.

\bibitem{IqbalGG16}
H.~Yasin, U.~Iqbal, B.~Kr{\"{u}}ger, A.~Weber, and J.~Gall, ``A dual-source
  approach for {3D} pose estimation from a single image,'' in \emph{CVPR},
  2016.

\bibitem{LinLLWC17}
M.~Lin, L.~Lin, X.~Liang, K.~Wang, and H.~Cheng, ``Recurrent {3D} pose sequence
  machines,'' in \emph{{CVPR}}, 2017.

\bibitem{AgarwalCVPR2004}
A.~Agarwal and B.~Triggs, ``{3D human pose from silhouettes by relevance vector
  regression},'' in \emph{CVPR}, 2004.

\bibitem{RogezRROT08}
G.~Rogez, J.~Rihan, S.~Ramalingam, C.~Orrite, and P.~H.~S. Torr, ``Randomized
  trees for human pose detection,'' in \emph{CVPR}, 2008.

\bibitem{SminchisescuCVPR2005}
C.~Sminchisescu, A.~Kanaujia, Z.~Li, and D.~N. Metaxas, ``Generative modeling
  for continuous non-linearly embedded visual inference,'' in \emph{CVPR},
  2005.

\bibitem{BoCVPR2008}
L.~Bo, C.~Sminchisescu, A.~Kanaujia, and D.~Metaxas, ``Fast algorithms for
  large scale conditional {3D} prediction,'' in \emph{CVPR}, 2008.

\bibitem{ShakhnarovichCVPR2003}
G.~Shakhnarovich, P.~Viola, and T.~Darrell, ``Fast pose estimation with
  parameter-sensitive hashing,'' in \emph{CVPR}, 2003.

\bibitem{LiZC15}
S.~Li, W.~Zhang, and A.~B. Chan, ``Maximum-margin structured learning with deep
  networks for {3D} human pose estimation,'' in \emph{ICCV}, 2015.

\bibitem{TekinBMVC2016}
B.~Tekin, I.~Katircioglu, M.~Salzmann, V.~Lepetit, and P.~Fua, ``Structured
  prediction of {3D} human pose with deep neural networks,'' in \emph{BMVC},
  2016.

\bibitem{PavlakosZDD17}
G.~Pavlakos, X.~Zhou, K.~G. Derpanis, and K.~Daniilidis., ``{ Coarse-to-fine
  volumetric prediction for single-image {3D} human pose},'' in \emph{CVPR},
  2017.

\bibitem{ZhouZLDD16}
X.~Zhou, M.~Zhu, S.~Leonardos, K.~Derpanis, and K.~Daniilidis, ``Sparseness
  meets deepness: {3D} human pose estimation from monocular video,'' in
  \emph{CVPR}, 2016.

\bibitem{SunSLW17}
X.~Sun, J.~Shang, S.~Liang, and Y.~Wei, ``Compositional human pose
  regression,'' in \emph{ICCV}, 2017.

\bibitem{Simo-SerraQTM13}
E.~Simo{-}Serra, A.~Quattoni, C.~Torras, and F.~Moreno{-}Noguer, ``A joint
  model for {2D} and {3D} pose estimation from a single image,'' in
  \emph{CVPR}, 2013.

\bibitem{ZhouT14}
F.~Zhou and F.~D. la~Torre, ``Spatio-temporal matching for human detection in
  video,'' in \emph{ECCV}, 2014.

\bibitem{TekingMSF17}
B.~Tekin, P.~Marquez-Neila, M.~Salzmann, and P.~Fua, ``Learning to fuse {2D}
  and {3D} image cues for monocular body pose estimation,'' in \emph{ICCV},
  2017.

\bibitem{TomeRA17}
D.~Tome, C.~Russell, and L.~Agapito, ``{Lifting from the deep: Convolutional
  {3D} pose estimation from a single image},'' in \emph{CVPR}, 2017.

\bibitem{MehtaRCFSXT17}
D.~Mehta, H.~Rhodin, D.~Casas, P.~Fua, O.~Sotnychenko, W.~Xu, and C.~Theobalt,
  ``Monocular {3D} human pose estimation in the wild using improved cnn
  supervision,'' in \emph{3DV}, 2017.

\bibitem{ZhouHSXW17}
X.~Zhou, Q.~Huang, X.~Sun, X.~Xue, and Y.~Wei, ``Towards {3D} human pose
  estimation in the wild: A weakly-supervised approach,'' in \emph{ICCV}, 2017.

\bibitem{DesouzaGCL17}
C.~R. de~Souza, A.~Gaidon, Y.~Cabon, and A.~Lopez, ``Procedural generation of
  videos to train deep action recognition networks,'' in \emph{CVPR}, 2017.

\bibitem{HuangR17}
S.~Huang and D.~Ramanan, ``Expecting the unexpected: Training detectors for
  unusual pedestrians with adversarial imposters,'' in \emph{CVPR}, 2017.

\bibitem{VarolRMMBLS17}
G.~Varol, J.~Romero, X.~Martin, N.~Mahmood, M.~Black, I.~Laptev, and C.~Schmid,
  ``{ Learning From Synthetic Humans},'' in \emph{CVPR}, 2017.

\bibitem{RogezS18}
G.~Rogez and C.~Schmid, ``Image-based synthesis for deep 3{D} human pose
  estimation,'' \emph{IJCV}, vol. 126, no.~9, pp. 993--1008, 2018.

\bibitem{Lassner0KBBG17}
C.~Lassner, J.~Romero, M.~Kiefel, F.~Bogo, M.~J. Black, and P.~V. Gehler,
  ``Unite the people: Closing the loop between {3D} and {2D} human
  representations,'' in \emph{{CVPR}}, 2017.

\bibitem{vgg}
K.~Simonyan and A.~Zisserman, ``Very deep convolutional networks for
  large-scale image recognition,'' in \emph{ICLR}, 2015.

\bibitem{deng2009imagenet}
J.~Deng, W.~Dong, R.~Socher, L.~Li, K.~Li, and L.~Fei-Fei, ``Imagenet: A
  large-scale hierarchical image database,'' in \emph{CVPR}, 2009.

\bibitem{andriluka14cvpr}
M.~Andriluka, L.~Pishchulin, P.~Gehler, and B.~Schiele, ``{2D} human pose
  estimation: New benchmark and state-of-the-art analysis,'' in \emph{CVPR},
  2014.

\bibitem{MehtaSMXSPT18}
D.~Mehta, O.~Sotnychenko, F.~Mueller, W.~Xu, S.~Sridhar, G.~Pons-Moll, and
  C.~Theobalt, ``Single-shot multi-person {3D} pose estimation from monocular
  {RGB},'' in \emph{3DV}, sep 2018.

\bibitem{KostrikovG14}
I.~Kostrikov and J.~Gall, ``Depth sweep regression forests for estimating {3D}
  human pose from images,'' in \emph{BMVC}, 2014.

\bibitem{LoperM0PB15}
M.~Loper, N.~Mahmood, J.~Romero, G.~Pons-Moll, and M.~J. Black, ``{SMPL}: A
  skinned multi-person linear model,'' \emph{ACM Trans. Graphics}, 2015.

\bibitem{BoS10}
L.~Bo and C.~Sminchisescu, ``Twin {G}aussian processes for structured
  prediction,'' \emph{IJCV}, 2010.

\bibitem{DWLHGWKG16}
Y.~Du, Y.~Wong, Y.~Liu, F.~Han, Y.~Gui, Z.~Wang, M.~Kankanhalli, and W.~Geng,
  ``Marker-less 3{D} human motion capture with monocular image sequence and
  height-maps,'' in \emph{ECCV}, 2016.

\bibitem{ZhouSZLW16}
X.~Zhou, X.~Sun, W.~Zhang, S.~Liang, and Y.~Wei, ``Deep kinematic pose
  regression,'' in \emph{ECCV Workshop}, 2016.

\bibitem{SanzariNP16}
M.~Sanzari, V.~Ntouskos, and F.~Pirri, ``Bayesian image based 3{D} pose
  estimation,'' in \emph{ECCV}, 2016.

\bibitem{KatirciogluTSLF18}
I.~Katircioglu, B.~Tekin, M.~Salzmann, V.~Lepetit, and P.~Fua, ``Learning
  latent representations of 3{D} human pose with deep neural networks,''
  \emph{IJCV}, 2018.

\bibitem{KinauerGCK17}
S.~Kinauer, R.~Guler, S.~Chandra, and I.~Kokkinos, ``Structured output
  prediction and learning for deep monocular {3D} human pose estimation,'' in
  \emph{{EMMCVPR}}, 2017.

\bibitem{LinMBHPRDZ14}
T.~Lin, M.~Maire, S.~J. Belongie, J.~Hays, P.~Perona, D.~Ramanan,
  P.~Doll{\'{a}}r, and C.~L. Zitnick, ``Microsoft {COCO:} common objects in
  context,'' in \emph{ECCV}, 2014.

\bibitem{HDM05}
M.~M\"{u}ller, T.~R\"{o}der, M.~Clausen, B.~Eberhardt, B.~Kr\"{u}ger, and
  A.~Weber, ``Documentation {M}o{C}ap database {HDM05},'' Universit\"{a}t Bonn,
  Tech. Rep. CG-2007-2, 2007.

\bibitem{YangLOLW17}
W.~Yang, S.~Li, W.~Ouyang, H.~Li, and X.~Wang, ``Learning feature pyramids for
  human pose estimation,'' in \emph{{ICCV}}, 2017.

\bibitem{ChuYOMYW17}
X.~Chu, W.~Yang, W.~Ouyang, C.~Ma, A.~L. Yuille, and X.~Wang, ``Multi-context
  attention for human pose estimation,'' in \emph{{CVPR}}, 2017.

\bibitem{GkioxariTJ16}
G.~Gkioxari, A.~Toshev, and N.~Jaitly, ``Chained predictions using
  convolutional neural networks,'' in \emph{{ECCV}}, 2016.

\bibitem{HuR16}
P.~Hu and D.~Ramanan, ``Bottom-up and top-down reasoning with hierarchical
  rectified gaussians,'' in \emph{{CVPR}}, 2016.

\bibitem{CarreiraAFM16}
J.~Carreira, P.~Agrawal, K.~Fragkiadaki, and J.~Malik, ``Human pose estimation
  with iterative error feedback,'' in \emph{{CVPR}}, 2016.

\bibitem{NewellHD17}
A.~Newell, Z.~Huang, and J.~Deng, ``Associative embedding: End-to-end learning
  for joint detection and grouping,'' in \emph{NIPS}, 2017.

\bibitem{InsafutdinovAPT17}
E.~Insafutdinov, M.~Andriluka, L.~Pishchulin, S.~Tang, E.~Levinkov, B.~Andres,
  and B.~Schiele, ``Arttrack: Articulated multi-person tracking in the wild,''
  in \emph{{CVPR}}, 2017.

\bibitem{InsafutdinovPAAS16}
E.~Insafutdinov, L.~Pishchulin, B.~Andres, M.~Andriluka, and B.~Schiele,
  ``Deepercut: A deeper, stronger, and faster multi-person pose estimation
  model,'' in \emph{ECCV}, 2016.

\bibitem{IqbalG16}
U.~Iqbal and J.~Gall, ``Multi-person pose estimation with local joint-to-person
  associations,'' in \emph{{ECCV} Workshops}, 2016.

\end{thebibliography}

\begin{IEEEbiography}[{\includegraphics[width=1in,height=1.25in,clip,keepaspectratio]{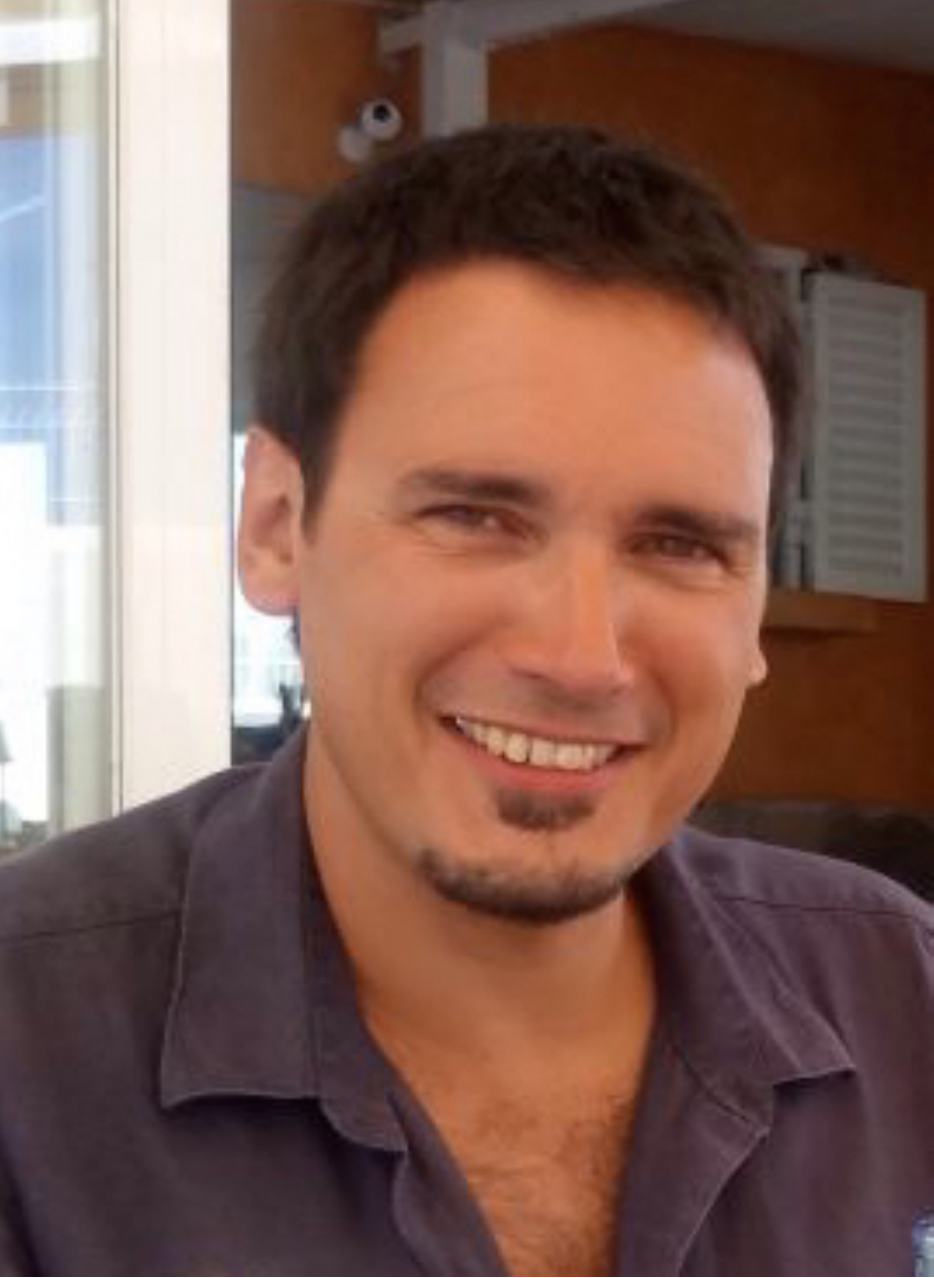}}]{Gr\'egory Rogez}
holds a M.Eng. in physics from the Ecole Nationale Sup\'erieure  de Physique de Marseille (now Centrale Marseille), France,  a M.Sc. degree in biomedical engineering and a Ph.D. degree in computer vision both from the University of Zaragoza, Spain.  Dr. Rogez was a visiting student (2007-2008) and research fellow (2009-2010) in the Computer Vision group of Oxford Brookes University. His work on monocular human body pose analysis received the best Ph.D. thesis award from the Spanish Association on Pattern Recognition (AERFAI) for the period 2011-2013.  Dr. Rogez was awarded a competitive Marie Curie Fellowship to visit the University of California between 2013 and 2015. He is currently a Research Scientist with the THOTH team at Inria Grenoble Rh\^one-Alpes.  His research interests include computer vision and machine learning,  with a special focus on understanding people from visual data, \ie, human detection, 2D/3D pose estimation,  action recognition and  object manipulation.
\end{IEEEbiography}

\begin{IEEEbiography}[{\includegraphics[width=1in,height=1.25in,clip,keepaspectratio]{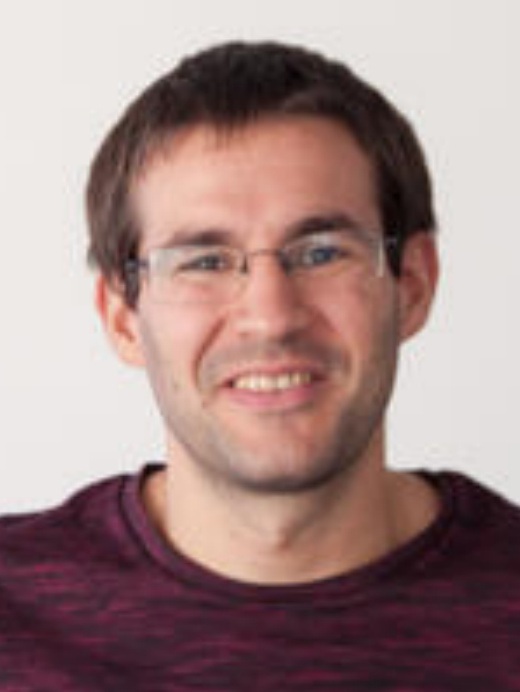}}]{Philippe Weinzaepfel}
received a M.Sc.  degree from Universit\'e Grenoble Alpes, France, and Ecole Normale Sup\'erieure de Cachan, France, in 2012.
He was a PhD student in the Thoth
team, at Inria Grenoble and LJK, from 2012 until 2016,
and received a PhD degree in computer science
from Universit\'e Grenoble Alpes  in 2016.
He is currently a Research Scientist at NAVER LABS Europe, France, in the computer vision group. 
His research interests include computer vision and machine learning, with special interest in video understanding and action recognition.
\end{IEEEbiography}

\begin{IEEEbiography}[{\includegraphics[width=1in,height=1.25in,clip,keepaspectratio]{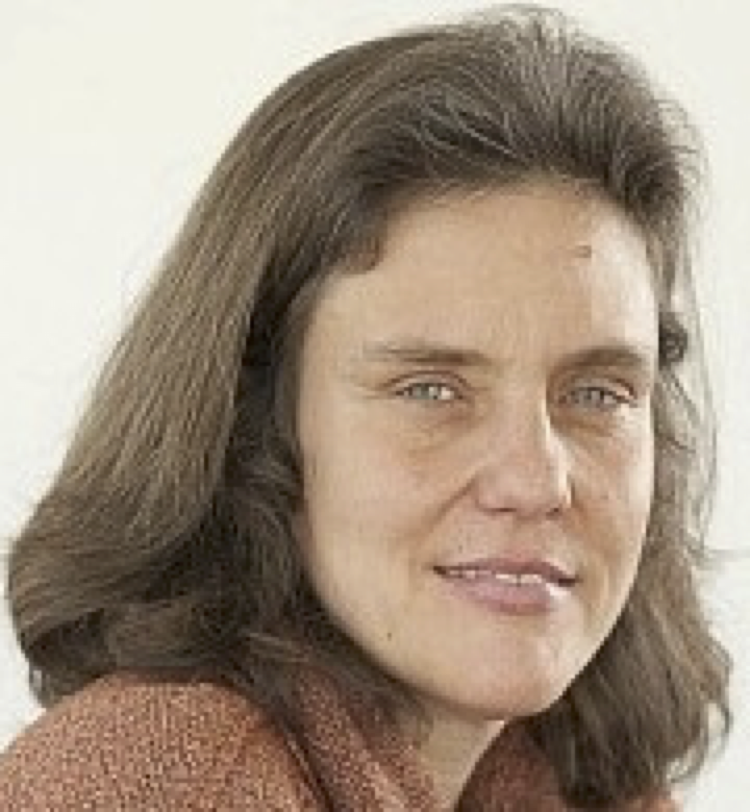}}]{Cordelia Schmid}
 holds a M.S. degree in computer science from the
University of Karlsruhe and a Doctorate, also in computer science,
from the Institut National Polytechnique de Grenoble. Her
doctoral thesis received the best thesis award from INPG in 1996. 
Dr. Schmid was a post-doctoral research assistant in the Robotics
Research Group of Oxford University in 1996--1997. Since 1997 she has
held a permanent research position at Inria Grenoble Rh\^one-Alpes,
where she is a Research Director. Dr. Schmid
has been an Associate Editor for IEEE PAMI (2001--2005) and for IJCV 
(2004--2012), editor-in-chief for IJCV (2013--2018), a program chair of
IEEE CVPR 2005 and ECCV 2012 as well as a general chair of IEEE CVPR
2015 and ECCV 2020. In 2006, 2014, 2016 and 2018, she was awarded 
the Longuet-Higgins resp. Koenderink prize for fundamental contributions in computer
vision that have withstood the test of time. She is a fellow of
IEEE. She was awarded an ERC advanced grant in 2013, the Humbolt
research award in 2015 and the Inria \& French Academy of Science
Grand Prix in 2016. She was elected to the German National
Academy of Sciences, Leopoldina, in 2017. 
\end{IEEEbiography}

\end{document}